\documentclass[journal]{IEEEtran}
\usepackage{graphicx}
\usepackage{amsmath}
\usepackage{rotating}
\usepackage{lineno}
\usepackage{subcaption}
\usepackage{threeparttable}
\usepackage[utf8]{inputenc}
\usepackage[hidelinks]{hyperref}
\usepackage{multirow}
\usepackage[english]{babel}
\usepackage{nomencl}
\makenomenclature 
\usepackage{booktabs}
\usepackage{tabu}
\usepackage{caption}
\usepackage{amssymb}
\usepackage{comment}
\usepackage{cite}
\usepackage{longtable}
\DeclareCaptionLabelFormat{andtable}{#1~#2  \& \tablename~\thetable}
\usepackage[linesnumbered,ruled,vlined]{algorithm2e}
\SetKwInput{KwInput}{Input}                
\SetKwInput{KwOutput}{Output}              
\usepackage{ltablex,array}

\hypersetup
{
 unicode=false,     
 pdftoolbar=true,    
 pdfmenubar=true,    
 pdffitwindow=false,   
 pdfstartview={FitH},  
 pdftitle={My title},  
 pdfauthor={Author},   
 pdfsubject={Subject},  
 pdfcreator={Creator},  
 pdfproducer={Producer}, 
 pdfkeywords={keywords}, 
 pdfnewwindow=true,   
 colorlinks=true,    
 linkcolor=red,     
 citecolor=blue,    
 filecolor=magenta,   
 urlcolor=cyan      
}

\ifCLASSINFOpdf
\else
\fi
\begin{document}
%
\title{A Structure-Aware Lane Graph Transformer Model for Vehicle Trajectory Prediction}

\author{Zhanbo~Sun, Caiyin~Dong, Ang Ji~\IEEEmembership{Member,~IEEE}, Ruibin~Zhao, and~Yu~Zhao
\thanks{The work is supported by the National Natural Science Foundation of China via grant 52072316 and 52302418, the Sichuan Science and Technology Program via grant 2024NSFSC0942, the Fundamental Research Funds for the Central Universities via grant 2682023CX047, and the Postdoctoral International Exchange Program via grant YJ20220311. \textit{(Corresponding author: Ang Ji)}

Zhanbo Sun, Caiyin Dong, Ang Ji, and Yu Zhao are with the School of Transportation and Logistics, and the National Engineering Laboratory of Integrated Transportation Big Data Application Technology at Southwest Jiaotong University, Chengdu, China (e-mail: zhanbo.sun@home.swjtu.edu.cn, cydong@my.swjtu.edu.cn, zy\_aureola@my.swjtu.edu.cn, ang.ji@swjtu.edu.cn).

Ruibin Zhao is with Zelos Intelligent Technology Limited Company, Suzhou, Jiangsu, China (email: z\_ruibin@163.com)


}}
\maketitle

\begin{abstract}
Accurate prediction of future trajectories for surrounding vehicles is vital for the safe operation of autonomous vehicles. This study proposes a Lane Graph Transformer (LGT) model with structure-aware capabilities. Its key contribution lies in encoding the map topology structure into the attention mechanism. To address variations in lane information from different directions, four Relative Positional Encoding (RPE) matrices are introduced to capture the local details of the map topology structure. Additionally, two Shortest Path Distance (SPD) matrices are employed to capture distance information between two accessible lanes. Numerical results indicate that the proposed LGT model achieves a significantly higher prediction performance on the Argoverse 2 dataset. Specifically, the minFDE$_6$ metric was decreased by 60.73\% compared to the Argoverse 2 baseline model (Nearest Neighbor) and the b-minFDE$_6$ metric was reduced by 2.65\% compared to the baseline LaneGCN model. Furthermore, ablation experiments demonstrated that the consideration of map topology structure led to a 4.24\% drop in the b-minFDE$_6$ metric, validating the effectiveness of this model.

\end{abstract}

\begin{IEEEkeywords}
Autonomous vehicles, trajectory prediction, high definition map, structure-aware, Transformer
\end{IEEEkeywords}

%

\section{Introduction}
In recent years, driven by advancements in emerging technologies such as 5G communication and artificial intelligence, a new era of intelligent transportation systems -- autonomous driving, has gradually gained prominence in public awareness\cite{molakis2017}. Trajectory prediction, which enables autonomous vehicles (agents) to proactively perceive the future movements of surrounding traffic participants, is widely-recognized to be crucial for the safety of autonomous driving\cite{QCnet2023,M2I2022,tiv-Survey2022}. Results from trajectory prediction also form the basis for early detection and intervention to prevent potential hazards\cite{xie2019mining,wang2024assessing}. 

\begin{figure}[h]
    \centering
    \includegraphics[width=0.5\textwidth]{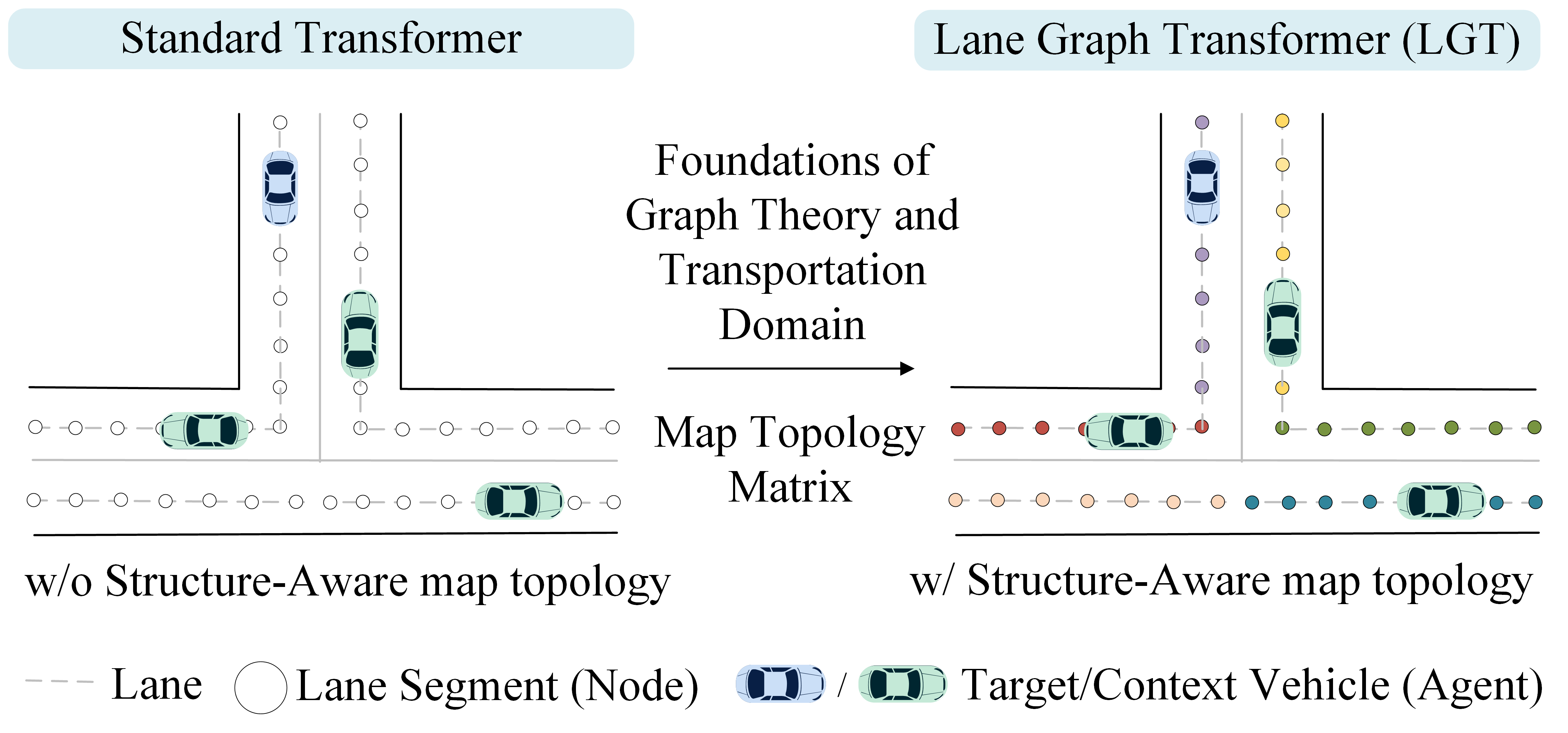}
    \caption{The critical issue addressed by the LGT model: compared to the standard Transformer, it integrates connectivity information between lanes, thus becoming structure-aware.}
    \label{fig:background}
\end{figure}

Predicting trajectories poses a significant challenge because of the complex interactions among traffic participants in dynamic scenarios. Historical trajectories of agents and high-definition (HD) maps provide valuable insights for understanding the context\cite{GNET2023, IPCC-TP2023}. Conveying environmental information (including surrounding agent trajectory data and HD map information) to the target vehicle can enable it to perceive the complex traffic environment. Hence, developing a model that can recognize agent-to-agent/map information is a pivotal element for precision in forecasting future trajectories. The interactions mainly manifest in the influence of the local map structure (agent-map interactions) and surrounding vehicles (agent-agent interactions) on the future intent of the target vehicle.

Existing trajectory prediction models with vectorization can be broadly classified into three categories\cite{laformer2023}: pooling-based methods\cite{SocialLSTM2016, SS-LSTM2018}, graph neural network (GNN) methods\cite{Social-Attention2018, tiv-mo2023, tiv-guo2023, tiv-Spatio-Temporal-Graph2023}, and attention-based methods\cite{Lane-transformer2023, tiv-zuo2023, tiv-Interaction-Aware2023, tivAI-TP2023}. The pooling-based method achieves interaction by directly combining information from neighboring entities and self-information. However, it struggles to effectively capture the dynamics of interactions in real driving scenarios\cite{wang2023wsip}. GNN methods, using a graph structure deliberately designed by humans, can adeptly capture relationships between nodes but grapple with the challenge of over-smoothing\cite{Rethinking-Graph2021}. Over the past few years, Transformer\cite{attention2023} has become a core component of many state-of-the-art machine learning models and is widely employed in trajectory prediction. Nevertheless, most prediction methods\cite{HiVT2022, Scene-Transformer2022, Wayformer2023, AgentFormer2021} rely on a standard Transformer where the self-attention for the subject node only accounts for the semantic similarity between the subject node and others, which disregards the structural information of nodes on the graph and the relationships between pairs of nodes. Consequently, the standard Transformer struggles to recognize map topology information, making it challenging to achieve effective lane graph representation. This raises the question: \textit{How can we integrate map topology into the Transformer structure?}

The integration of graph structures into the Transformer architecture has been extensively explored in the field of graph theory\cite{Directed-Graphs2023, Bad-for-Graph2021, Not-be-Powerful2022}, which can be classified into three categories: (i) GNNs as Auxiliary Modules, (ii) Improved Positional Embedding from Graphs, and (iii) Improved Attention Matrix from Graphs\cite{Transformer-for-Graphs2022}. Evidence\cite{Transformer-for-Graphs2022, Biconnectivity2023} suggests that Improved Attention Matrix from Graphs, which injects graph priors into the attention mechanism through graph bias terms, has shown superior performance. In the field of trajectory prediction, extensive efforts have been made in designing map topology structures as graphs\cite{LaneGCN2020, GNET2023, VectorNet2020}. For instance, Vectornet\cite{VectorNet2020} creates a global interaction graph that includes agents and lanes. Meanwhile, LaneGCN\cite{LaneGCN2020} designs a graph network with lane segments as nodes and incorporates connections between lanes, thereby enhancing the overall graph structure. Therefore, applying graph theory-based methods to trajectory prediction is a feasible improvement.

Motivated by these findings, we propose an LGT-based model with a novel lane graph representation in trajectory prediction. The model design aims to address the key issues depicted in \autoref{fig:background}, specifically by endowing the model with structural awareness capabilities. It integrates the Relative Positional Encoding (RPE) and the Shortest Path Distance (SPD) matrices of the map topology structure as bias matrices into the attention mechanism. This approach is applied to the lane graph representation in trajectory prediction, thereby addressing gaps in existing research on trajectory prediction based on attention mechanisms. The main contributions of this study include:

\begin{itemize}
    \item Leveraging domain-specific knowledge in traffic systems, we propose an LGT structure that integrates the map topology information into the Transformer, substantiated by graph theory. The experimental results show that this approach markedly improves the accuracy of trajectory prediction compared to benchmark models.
    \item We construct RPE and SPD matrices to represent the map topology information and integrate them into the attention mechanism. The results demonstrate that both matrices can enhance the prediction accuracy, and the RPE matrices exhibit more significant effects.
    \item   We employed a local attention mechanism focusing on a specific number of neighbors to facilitate interaction between agents and maps. The optimal number of neighbors in different contexts is sought through grid search, which can hopefully provide statistical references for future related studies.
\end{itemize}

This paper is organized as follows. \hyperref[sec:lit]{Section~\ref*{sec:lit}} provides a summary of previous research work. \hyperref[sec:method]{Section~\ref*{sec:method}} introduces the architecture of the proposed model. In \hyperref[sec:results]{Section~\ref*{sec:results}}, we conduct experiments and present the results. Finally, \hyperref[sec:conclusion]{Section~\ref*{sec:conclusion}} concludes the findings and outlines future research directions. 

\section{Related Work}
\label{sec:lit}
In this section, we categorize existing trajectory prediction models into three classes. Also, we provide a detailed overview of methods for encoding graph structures in graph research.

\textbf{Pooling-based Strategy.} Social Long Short-Term Memory (LSTM)\cite{SocialLSTM2016} and Social-Scene-LSTM\cite{SS-LSTM2018} model the trajectory of each individual agent through a separate network, employing a Social pooling layer to receive and merge hidden state information from surrounding neighbors. However, this method relies on fixed pooling, lacking the ability to consider the impact of local neighbors on pedestrian movement\cite{Social-Attention2018}. 

\textbf{GNN-based Strategy.} Compared to the fixed operations of the pooling approach, the GNN-based method offers several advantages: (i) Adaptability to the heterogeneity (vehicles, pedestrians, etc.) of traffic scenes; (ii) Improved maintenance of dynamic interaction models through gradual and iterative updates of node features; (iii) Enhanced intuitiveness in interaction. VectorNet\cite{VectorNet2020} encodes map elements and agent trajectories and then deploys a global interaction graph to fuse map and agent features. However, the vanilla graph network with undirected full connections faces challenges in representing real traffic scenarios. Various improvements have been implemented by optimizing the graph net structure: Social Attention considers the importance of different pedestrians\cite{Social-Attention2018}; LaneGCN examines the types of connections between lanes\cite{LaneGCN2020}; Mo et al. constructed heterogeneous graphs incorporating edge characteristics\cite{Mo2022}. However, the construction of traffic scene graphs using GNNs inherently encounters the following problems: (i) information about non-neighboring nodes needs to be aggregated multiple times, (ii) the over-smoothing problem due to repeated local aggregation\cite{Rethinking-Graph2021}.

\textbf{Attention-based Strategy.} Transformers have been widely adopted in trajectory prediction due to their strong capabilities in modeling long sequences and directly attending to nodes in map networks from the global perspective. For example, MmTransformer\cite{Multimodal2021} comprises three independent stacked Transformer models, which aggregate historical trajectories, map information, and interaction information. The Multi-modal Transformer\cite{Multi-modal2021}, leveraging 1D convolutional layers and LSTM layers, encodes features of both agents and maps. It then employs standard Transformers to manage agent-agent interaction and agent-map interaction, followed by concatenating these interactive features and feeding them into the decoder for future trajectory prediction. Similar to these approaches, recent studies with standard Transformers do not emphasize the structural information of map graphs, hindering effective lane graph representation\cite{Multi-Head2020, Multi-modal2021, HiVT2022, Scene-Transformer2022}. Simon et al.\cite{Directed-Graphs2023} emphasize the importance of the attention mechanism to recognize graph structures.

\begin{figure*}[!t]
\centering
\includegraphics[width=1\linewidth]{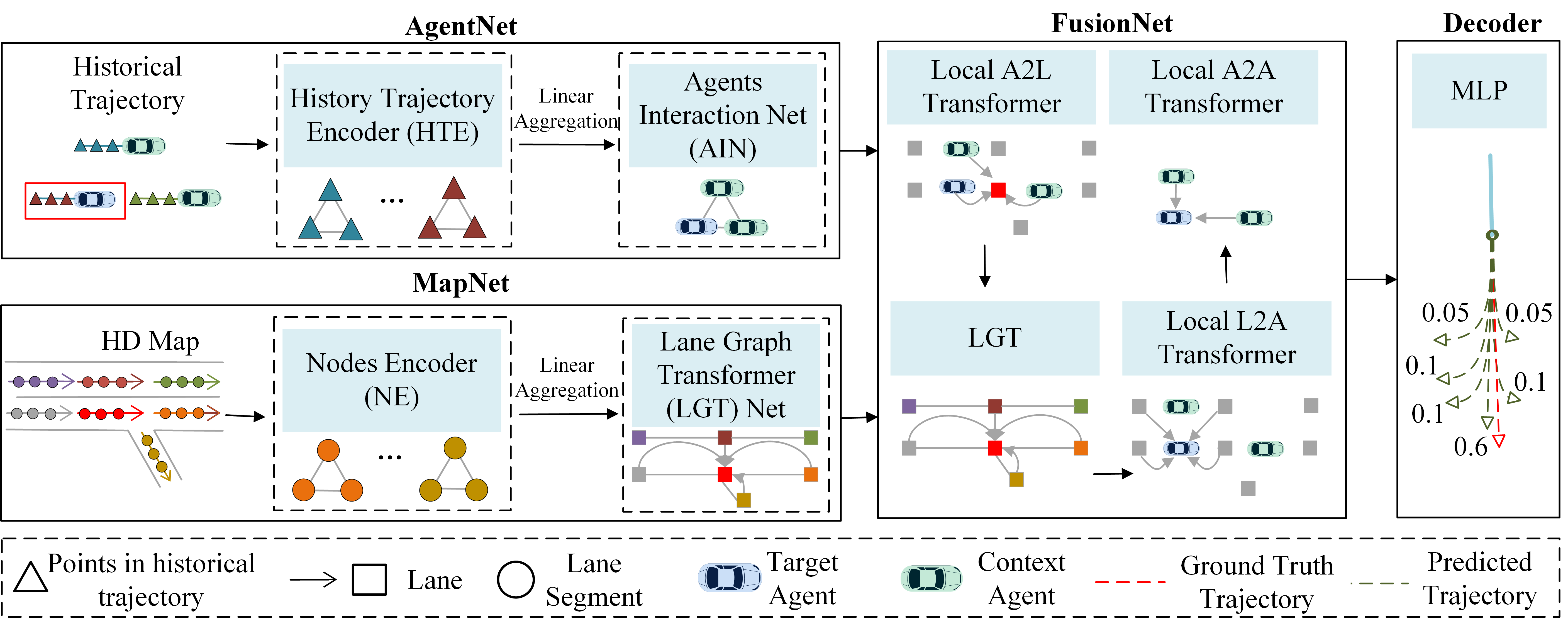}
\label{fig:framework}
\caption{The overall framework of the proposed model, including AgentNet, MapNet, FusionNet, and Decoder.\label{fig:overall}}
\end{figure*}

\textbf{Graph Theory Applications in Transformers.} In recent years, numerous variants of Transformers have emerged to accommodate graph-structured data, drawing from insights in graph theory\cite{Biconnectivity2023, generalization-dwivedi2021, GRPE2022}. Graphormer\cite{Bad-for-Graph2021} encodes graph information into the Transformer architecture by incorporating centrality, spatial, and edge encoding. Universal Relative Positional Encoding (URPE)\cite{Not-be-Powerful2022} theoretically proves that the external dot product of Softmax with a Toeplitz matrix can achieve universal approximation. Graphormer-GD\cite{Biconnectivity2023}, utilizing the SPD matrix, contributes to recognizing the bi-connectivity of a graph and attains optimal performance. In the field of graphs, incorporating structural information from graphs into Transformer architectures has demonstrated remarkable effectiveness. Therefore, it is promising to instill Transformer models with structural awareness using this approach. 

In summary, attention-based approaches have demonstrated satisfactory performance in trajectory prediction compared to other interaction methods. However, existing attention-based interaction methods directly connect nodes, overlooking the map topology structure information. Fortunately, research in the field of graph theory has effectively embedded graph structures into attention mechanisms, yielding excellent results. Therefore, the focal point of this paper is to integrate relevant methods from the graph theory domain, along with incorporating specialized knowledge from the transportation field, to explore how to integrate the map topology structure into the attention mechanism for enhanced prediction accuracy.

\section{Methodology}
\label{sec:method}
This section elucidates our LGT framework in a pipelined manner. The proposed LGT architecture is illustrated in \autoref{fig:overall}, with each module within this framework being pluggable. The model comprises four key components: (a) AgentNet includes two parts: History Trajectory Encoder (HTE), facilitating the encoding of a single agent's trajectory features at different historical moments, and Agents Interaction Net (AIN), managing interactions among agents within the scene. (b) Leveraging existing research experience\cite{LaneGCN2020, VectorNet2020}, we segment each road into multiple lanes, and within each lane, there are several lane segments (nodes), as depicted in \autoref{fig:background}. This partitioning gives rise to two components of MapNet: Nodes Encoder (NE), facilitating the encoding of features among nodes on the same lane, and LGT managing interactions across different lanes. (c) FusionNet employs a sequentially stacked transformer structure to achieve interactions between Agents to Lanes (A2L), Lanes to Lanes (L2L), Lanes to Agents (L2A), and Agents to Agents (A2A). (d) A Multilayer Perceptron (MLP)-based model generates future trajectories of agents. A comprehensive overview of the components is provided below. 

\subsection{Preliminaries}
In this section, we recap the preliminaries in Transformer. The Transformer architecture is a composition of Transformer layers\cite{attention2023}. Each Transformer layer has two parts: an attention module and a position-wise feed-forward network (FFN). Let \textit{H} be the input to the attention module. The input \textit{H} is projected by three matrices $W_{Q}$, $W_{K}$, and $W_{V}$ to the corresponding representations $Q$, $K$, and $V$. The attention is then calculated as:

\begin{equation}
Q=HW_{Q} , K=HW_{K}, V=HW_{V} 
\end{equation}
\begin{equation}
A=\frac{QK^{T}}{\sqrt{d_{K}}} , \mathrm{Attention} (Q,K,V)=\mathrm{softmax} (A)V
\end{equation}

\noindent where \textit{A} is a matrix capturing the similarity between queries and keys, $d_{K}$ represents the dimensionality of \textit{K}.

\begin{figure*}[!t]
\centering
\includegraphics[width=1\linewidth]{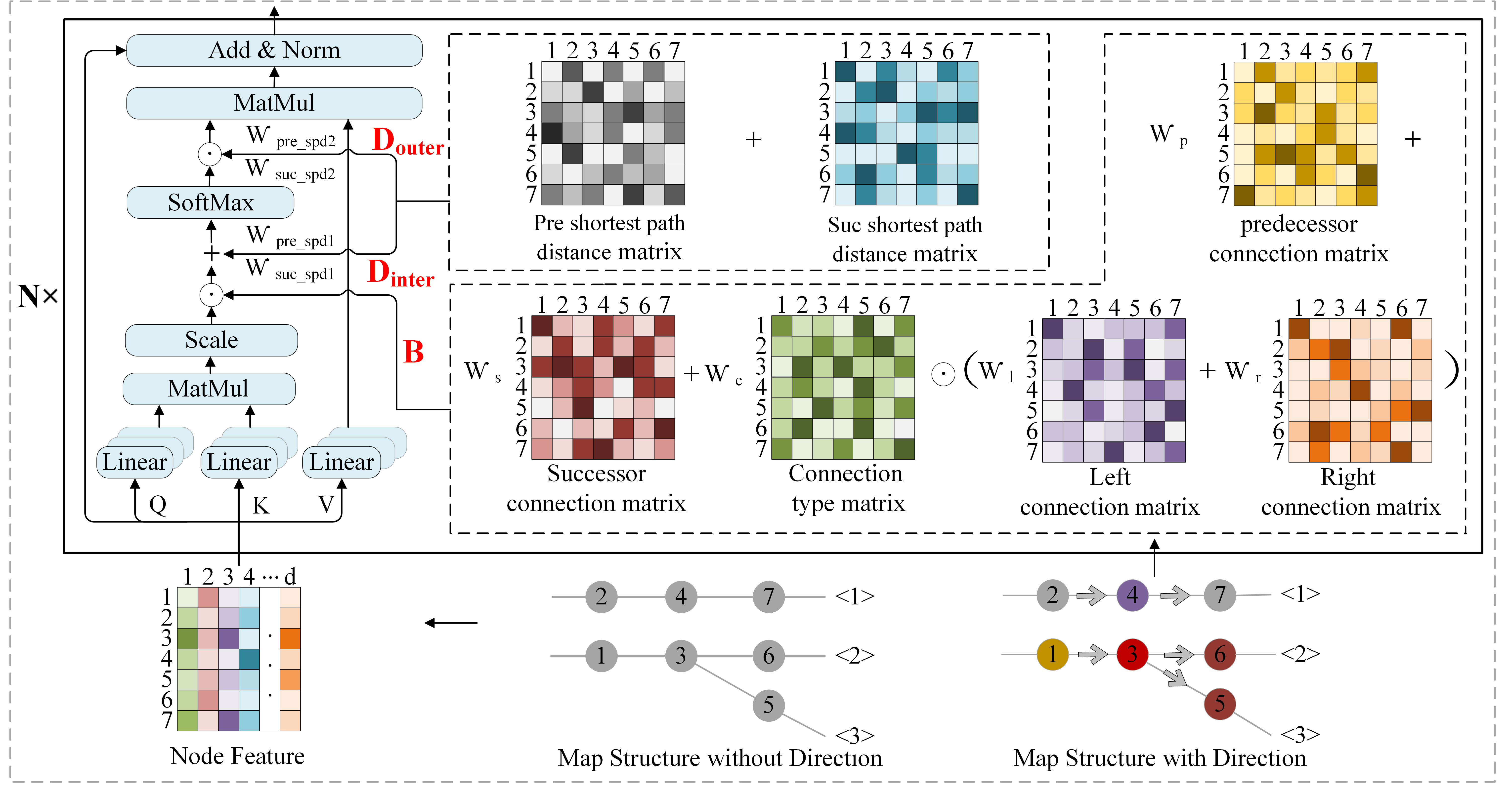}
\label{fig:LGT}
\caption{An illustration of the LGT framework. It embeds map topology information into the attention matrix by introducing three bias matrices \textit{B}, $D_{inter}$, $D_{outer}$ (highlighted in red font). Noteworthy that $D_{inter}$ and $D_{outer}$ exhibit a similar structure but with distinct weights.\label{fig:LGT1}}
\end{figure*}

\subsection{Model Input and Output}
The task of trajectory prediction involves utilizing historical trajectories of the target agents and their surrounding agents for \textit{T} time steps and the HD map information to predict \textit{K} potential future trajectories of target agents for \textit{T'} time steps. Assuming a scenario with $N_{a}$ agents, we employ an agent-centric strategy\cite{DenseTNT2021, TNT2020, MultiPath++2021} that normalizes all inputs to a coordinate system centered around the target agent. The origin is set at the position of the target agent when $t=0$. Regarding the HD map information, the input includes the centerline coordinates, lane ID, and lane type (e.g., Vehicle lane, Bus lane, etc). For the \textit{i-th} agent, the historical state $P_{i}$ is represented as follows:

\begin{equation}
P_{i} =[P_{i}^{-T+1}, P_{i}^{-T+2}, \cdots , P_{i}^{0}] 
\end{equation}

Here $P_{i}^{t}$ denotes the state of agent \textit{i} at time step \textit{t} and is composed as follows:

\begin{equation}
P_{i}^{t}=[x_{i}^{t},y_{i}^{t},padding_{i}^{t}, category_{i}^{t}, type_{i}^{t}, h_{i}^{t}, v_{i,x}^{t}, v_{i,y}^{t}] 
\end{equation}

\noindent where $x_{i}^{t},y_{i}^{t}$ denotes coordinates of the agent’s center; $padding_{i}^{t}$ indicates whether the agent lacks historical trajectory at moment \textit{t} (filled with 0 if absent); $category_{i}^{t}$ represents the trajectory quality (from 0 to 3), where a higher value indicating more comprehensive historical trajectory; $type_{i}^{t}$ specifies the type of agent, e.g., vehicles, buses, pedestrians, etc; $h_{i}^{t}$ denotes the heading angle associated with the agent; and $v_{i,x}^{t}, v_{i,y}^{t}$ are the velocity of agent on X, Y axis. 

The output consists of the \textit{K} potential future trajectories of the target agents, where the \textit{k-th} future trajectory of agent \textit{i} can be represented as follows:

\begin{equation}
P_{i,k}=[P_{i,k}^{1}, P_{i,k}^{2},\cdots ,P_{i,k}^{T'}]
\end{equation}

\noindent where $P_{i,k}^{t}=[x_{i,k}^{t},y_{i,k}^{t}]$ denotes the coordinates of future trajectories.

\subsection{Agent Net}
\subsubsection{History Trajectory Encoder (HTE)}
The influence of historical trajectories on future predicted ones differs across various timestamps. Hence, we employ HTE to encode the temporal features of a single agent's trajectory. The historical states of $N_{a}$ agents are represented as $A_{input}\in\mathbb{R}^{N_{a}\times T \times M_{a}}$, in which $M_{a}$ is the number of input features. 

First, we utilize an MLP to transform the input features into $A_{t} \in \mathbb{R}^{N_{a}\times T \times D_{\mathrm{agent} }}$.

\begin{equation}
A_{t}=\mathrm{MLP} (A_{input)}
\end{equation}

\noindent where $\mathrm{MLP} (\cdot )$ denotes a multilayer perceptron network and $D_{\mathrm{agent}}$ represents the dimensionality of the transformed features.

Subsequently, a stacked transformer structure is employed to encode the features within \textit{T} time steps of an agent. The attention mechanism for the \textit{j-th} layer of the Transformer structure can be expressed as follows:

\begin{equation}
A_{a-t}^{j}=\mathrm{Attention} (Q=A_{a-t}^{j-1},K=A_{a-t}^{j-1},V=A_{a-t}^{j-1})
\end{equation}

\noindent where $A_{a-t}^{0}=A_{t}$, and $A_{a-t}^{j} \in \mathbb{R}^{N_{a}\times T \times D_{\mathrm{agent} }}$.

Finally, utilizing an MLP-based aggregator, a singular feature vector is constructed for each trajectory, resulting in the final output feature $A_{\mathrm{HTE}} \in \mathbb{R}^{N_{a}\times D_{\mathrm{agent} }}$.

\subsubsection{Agents Interaction Net (AIN)}
We implement the AIN module to identify differences among agents' trajectories and effectively gather information about surrounding neighbors. AIN also employs a Transformer structure, which is presented as follows:

\begin{equation}
A_{\mathrm{AIN} }^{j}=\mathrm{Attention} (Q=A_{\mathrm{AIN}}^{j-1},K=A_{\mathrm{AIN}}^{j-1},V=A_{\mathrm{AIN}}^{j-1})
\end{equation}

\noindent where $A_{\mathrm{AIN}}^{0}=A_{\mathrm{HTE}}$, and the final output features of $A_{\mathrm{AIN}} \in \mathbb{R}^{N_{a}\times D_{\mathrm{agent} }}$.

\subsection{Map Net}
\subsubsection{Nodes Encoder (NE)}
The NE is introduced to encode the features of nodes on the same lane, further aggregating these nodes into lane-level features. NE takes feature $M_{input} \in \mathbb{R}^{N_{l}\times N_{ls}\times M_{m}}$ as the input and computes the aggregated feature $ M_{\mathrm{NE} } \in \mathbb{R}^{N_{l}\times D_{\mathrm{map}}}$. $N_{l}$ represents the number of lanes in the scene, $N_{ls}$ is the number of nodes in each lane, $M_{m}$ denotes the dimensionality of inputs for the map, and $D_{\mathrm{map}}$ indicates the output dimension produced by the NE network. NE and HTE are networks with identical structures and data processing details but with different hyperparameters and weights. 

\subsubsection{Lane Graph Transformer (LGT) Net}
In each Transformer layer, each node can attend to the information at any position and process its representation. However, this operation faces a challenge in which the model needs to explicitly specify different positions or encode positional dependencies, such as locality, in the layers. For sequential data, one can either provide each position with an embedding (i.e., absolute positional encoding\cite{attention2023}) as input or encode the relative distance of any two positions (i.e., relative positional encoding\cite{Limits-of-Transfer2023, selfattentionrpe2018}) in the Transformer layer.

However, compared to one-dimensional sequential data, map graphs display complex connection patterns, encompassing differences in spatial distances, various types of node connections, etc. Traditional encoding methods often struggle to capture these intricate relationships. In recent years, efforts have been spent on positional encoding in graph theory. Inspired by the concepts presented in Graphormer\cite{Bad-for-Graph2021} and Graphormer-GD\cite{Biconnectivity2023}, we design a novel LGT framework, which is depicted in \autoref{fig:LGT1}.

The attention mechanism within the LGT model is designed as follows:

\begin{equation}
\begin{aligned}
&\mathrm{LGTAttn} (Q,K,V,B,D_{\text{inter}},D_{\text{outer}})= [D_{\text{outer}} \odot \\
& \mathrm{softmax} (\frac{QK^{T}}{\sqrt{d_{K}}} \odot B + D_{\text{inter}})]V \\
&B=W_{p}M_{p}+W_{s}M_{s}+W_{c}M_{c}(W_{l}M_{l}+W_{r}M_{r}) \\
&D_{inter}=W_{pre\_spd1}M_{pre\_spd}+W_{suc\_spd1}M_{suc\_spd} \\
&D_{outer}=W_{pre\_spd2}M_{pre\_spd}+W_{suc\_spd2}M_{suc\_spd}
\end{aligned}
\end{equation}

\noindent where $M_{p}, M_{s}, M_{l}, M_{r}, M_{c}$ represent the matrices for the predecessor, successor, left, and right connections, and the connection type matrix of the map topology structure. $M_{pre\_spd}, M_{suc\_spd}$ represent the SPD matrices for the predecessor and successor connections. $W$ is the weight corresponding to each matrix.

\textbf{RPE Matrices:} We employ bias matrices \textit{B} to represent the spatial relativity of the map topology. Since vehicles' longitudinal speeds are typically higher than their lateral speeds, agents demonstrate more dependencies in the preceding and succeeding neighbors. To capture distinctions in directional traffic movements, we define four connectivity relations in the same manner as LaneGCN\cite{LaneGCN2020}: predecessor, successor, left neighbor, and right neighbor. Utilizing these four connectivity relations, we construct four RPE matrices ($M_{p}$, $M_{s}$, $M_{l}$, $M_{r}$) to represent the local details of the map topology. 

In addition, lateral connectivity can impact lane-changing, e.g., double yellow lines prohibit lane changes. Hence, we construct a connection type matrix, denoted as matrix $M_{c}$, and perform matrix multiplication to indicate the lateral connectivity between adjacent lanes, represented by $W_{c}M_{c}(W_{l}M_{l}+W_{r}M_{r})$.

\textbf{SPD Matrices:} The SPD matrices represent logical distances between any two accessible lanes. For any lane \textit{C}, we define two matrices PRE\_SPD and SUC\_SPD that travel to or from \textit{C} respectively, which distinguish the SPD from the preceding and successive directions.

Unlike typical bi-directional graph structures, lanes in the map topology structure are uni-directional (vehicles cannot travel against the flow). That is, only accessible lane nodes within the same direction are interconnected. Leveraging these characteristics, we design a sequentially expanding method that calculates the shortest paths starting from the origin node. During the expansion, reachable neighboring nodes are searched. The detailed generation process of PRE\_SPD is presented in \hyperref[algo:SPD]{Algorithm~\ref*{algo:SPD}}.

\begin{algorithm}
\caption{Generating PRE\_SPD matrix\label{algo:SPD}}
\KwIn{Pairs of reachable predecessor lanes $P$, the number of lanes $n$}
\KwOut{PRE\_SPD matrix}
\LinesNumbered
\For{$i=1$ to $n$}{
    Initialize an $n\times n$ PRE\_SPD matrix with all zeros\;
    $pre\_spd \leftarrow 1$, $neighbors \leftarrow [i]$, $appear\_lanes \leftarrow [i]$\;
    \While{$neighbors$ is not empty}{
        $neighbor \leftarrow []$\;
        \For{$j$ in $neighbors$}{
            Check if $j$ exists in $P$; if so, add the reachable predecessor lane ID to $neighbor$\;
        }
        Remove elements from $neighbor$ that already exist in $appear\_lanes$\;
        Add the lanes from $neighbor$ to $appear\_lanes$\;
        Fill PRE\_SPD matrix for lanes from $i$ to the $neighbor$ with $pre\_spd$\;
        $pre\_spd \leftarrow pre\_spd + 1$\;
        \If{$neighbors$ not empty}{
            $neighbors \leftarrow neighbor$\;
        }
        \Else{
            end\;
        }
    }
end
}
\end{algorithm}

\textbf{Details of matrix construction:} Typically, the nearest neighboring lanes to the target lane are considered more significant and assigned great values in the structural matrices (predecessor, successor, left, right, and SPD matrices). Therefore, we populate the reciprocal of the actual distance into the respective connection matrices with unfilled connections setting to 0. As for the connection type matrix, we apply one-hot encoding to convert connection types into binary vectors. Each category is denoted by a unique binary number, where 1 signifies its presence and the rest are set to 0. Subsequently, a straightforward MLP is employed to aggregate connection type features, which are then added to the matrices.

The LGT structure is constructed with stacked Transformer layers embedding topology information from maps. All layers in this structure share RPE and SPD matrices. Taking the output of NE as the input for LGT, we obtain the map information $M_{\mathrm{LGT} } \in \mathbb{R}^{N_{l}\times D}$, where D denotes the dimension of the outputs ($D_{\mathrm{agent}}=D_{\mathrm{map}}=D$). The attention matrices for the \textit{j-th} layer can be expressed as follows:

\begin{equation}
\begin{aligned}
M_{\mathrm{LGT}}^{j} &= \mathrm{LGTAttn} (Q=M_{\mathrm{LGT}}^{j-1}, K=M_{\mathrm{LGT}}^{j-1},\\
&\quad V=M_{\mathrm{LGT}}^{j-1}, \textit{B}, D_{inter}, D_{outer})
\end{aligned}
\end{equation}

\subsection{Fusion Net}
The Fusion Net is designed to amplify information interaction between each agent and the map, allowing the target agent to perceive its surroundings comprehensively. Drawing inspiration from the LaneGCN model, the interaction sequence unfolds as A2L, L2L, L2A, and A2A. Interacting with every object may not be necessary in a vast traffic scenario. Additionally, global interactions could require more computer memory\cite{Global-Intention2023}. Proximity to objects has a more pronounced impact on the agents' future trajectories, while information from distant objects can be processed through stacked transformers for secondary interaction. Therefore, we implement A2L, L2A, and A2A interactions using local attention\cite{Global-Intention2023, GANet2023}. The formula for the \textit{j-th} layer of local attention is expressed as follows:

\begin{equation}
\begin{aligned}
G^{j} &= \mathrm{Attention} (Q=G^{j-1}, K=e(G^{j-1}), \\
&\quad V=e(G^{j-1}))
\end{aligned}
\end{equation}

\noindent where $e(\cdot)$ denotes the \textit{e}-nearest neighbor algorithm that finds the \textit{e} nearest neighbors for each query.

Then, we exploit the LGT structure to execute L2L, with the final output features of Fusion Net denoted as $A_{\mathrm{encoder} } \in \mathbb{R}^{N_{a}\times D}$. Finally, this feature will serve as the input for subsequent decoding network components.

\subsection{Prediction Header}
The Fusion Net produces features that are then fed into the Prediction Header to output the final predicted trajectory. Each target agent predicts \textit{K} potential future trajectories, accompanied by confidence scores for each trajectory. To achieve this, we employ an MLP-based approach~\cite{ssllanes2022, tiv-intrinsic-interaction2023} as the Prediction Header. Due to the multimodality of trajectories, each trajectory prediction head uses the same structure but with different parameters.

\subsection{Training Loss}
For an agent with \textit{K} potential trajectories, we define the trajectory closest to the ground-truth trajectory at the final time step as the best-predicted trajectory, annotated $\hat{k}$.

We train the model with a combined loss function, incorporating regression, classification, and goal loss, which is formulated as follows:

\begin{equation}
\mathcal{L} = \mathcal{L}_{reg} +\mathcal{L}_{cls}+\mathcal{L}_{goal}
\end{equation}

For classification loss, the loss function is defined as:
\begin{equation}
\mathcal{L}_{cls}=\frac{1}{N(K-1)}  {\textstyle \sum_{i=1}^{N}}  {\textstyle \sum_{k \neq \hat{k}}}\mathrm{max} (0, c_{i}^{k} + \varepsilon - c_{i}^{\hat{k} })
\end{equation}

\noindent where \textit{N} is the number of agents and $c_{i}^{\hat{k}}$ represents the confidence score of the $k^{\mathrm{th}}$ predicted trajectory for agent \textit{i}. 

For regression loss, we apply the smooth L2 loss on all predicted steps of the positive trajectories:

\begin{equation}
\mathcal{L}_{reg}=\frac{1}{NT}  {\textstyle \sum_{i=1}^{N}}  {\textstyle \sum_{t=1}^{T}} reg(P_{i,\hat{k}}^{t} - P_{i,k^{\ast}}^{t}) 
\end{equation}

$P_{i,k^{\ast}}^{t}$ denotes the coordinates of the ground truth trajectory for agent \textit{i}.

We designate the predicted trajectory position at time $T$ as the goal. To underscore the significance of goal-achieving, we exploit the goal loss to penalize instances where the predicted position deviates significantly from the true position:
\begin{equation}
\mathcal{L}_{goal}=\frac{1}{N}  {\textstyle \sum_{i=1}^{N}}  reg(P_{i,\hat{k}}^{T'} - P_{i,k^{\ast}}^{T'}) 
\end{equation}

\section{Experiment and Numerical Results}
\label{sec:results}
In this section, we begin by introducing the dataset and evaluation metrics. Subsequently, the results of the proposed model are compared to those of state-of-the-art trajectory prediction models. Following that, ablation experiments are conducted to validate the effectiveness of the proposed LGT modules. Finally, the prediction results are examined and illustrated.

\begin{table*}
\caption{Results on the Argoverse 2 test set motion forecasting leaderboard\cite{leaderboard}, ranked by b-minFDE$_6$. The best results are highlighted in \textbf{bold} for all metrics.\label{tab:quantitatives}}
\centering
\resizebox{0.9\hsize}{!}{$
\begin{tabular}{l|cccc|ccc}
\toprule
Method & b-minFDE$_{6}\downarrow$ & minADE$_{6}\downarrow$ & minFDE$_{6}\downarrow$ & MR$_{6}\downarrow$ & minADE$_{1}\downarrow$ & minFDE$_{1}\downarrow$ & MR$_{1}\downarrow$ \\
\midrule
Argoverse 2 Baseline (NN) & - & 2.18 & 4.94 & 0.60 & 4.46 & 11.71 & 0.81 \\
\midrule
fsq & 3.11 & 1.11 & 2.50 & 0.40 & 2.75 & 7.38 & 0.73 \\
drivingfree & 3.03 & 1.17 & 2.58 & 0.49 & 2.47 & 6.26 & 0.72 \\
LGU & 2.77 & 1.05 & 2.15 & 0.37 & 2.77 & 6.91 & 0.73 \\
LaneGCN & 2.64 & \textbf{0.91} & 1.96 & \textbf{0.30} & 2.43 & 6.51 & 0.71 \\
\textcolor{red}{\textbf{LGT}} & \textbf{2.57} & 0.95 & \textbf{1.94} & \textbf{0.30} & \textbf{2.39} & \textbf{6.19} & \textbf{0.69} \\
\bottomrule
\end{tabular}
$}
\end{table*}

\begin{table*}[htbp]
\centering
\caption{Ablation experiments of LGT components and local attention. The experimental results are based on the Argoverse 2 validation set.\label{tab:ablation}}
\centering
\resizebox{0.7\hsize}{!}{$
\begin{tabular}{cccc|cccc}
\toprule
B & Dinter & Douter & Local Attention & b-minFDE$_6\downarrow$ & minADE$_6\downarrow$ & minFDE$_6\downarrow$ & MR$_6\downarrow$ \\
\midrule
 &  &  &  & 2.83 & 1.11 & 2.21 & 0.35 \\
\checkmark &  &  &  & 2.74 & 1.09 & 2.13 & 0.34 \\
\checkmark & \checkmark & \checkmark &  & 2.71 & 1.08 & 2.09 & 0.34 \\
\checkmark & \checkmark & \checkmark & \checkmark & \textbf{2.56} & \textbf{1.02} & \textbf{1.93} & \textbf{0.30} \\
\bottomrule
\end{tabular}
$}
\end{table*}

\subsection{Experiment Settings}
\textbf{Datasets.} We used the large-scale motion prediction dataset Argoverse 2\cite{Argoverse22023} to test the proposed method. The Argoverse 2 dataset contains 250, 000 scenarios spanning six cities. Each scenario contains historical trajectory data for various traffic participants, such as vehicles, buses, bicycles, and pedestrians, including their positions, velocities, heading angles, etc. Furthermore, it incorporates map structural details like lane centerlines, lane connectivity, and lane types. These scenarios were divided into three parts, training, validation, and testing data, with 199,908, 24,988, and 24,984 samples, respectively. Each scenario included 11 seconds of historical trajectories and contained the 2D, birds-eye-view centroid and heading of each tracked object sampled at 10 Hz. The first five seconds were used as input to the model, and the last six seconds were regarded as ground truth to evaluate the model performance.

\textbf{Metrics.} According to the evaluation criteria of the Argoverse 2 dataset\cite{Argoverse22023}, we evaluated the performance using the following metrics: minimum Average Displacement Error (minADE$_K$), minimum Final Displacement Error (minFDE$_K$), Miss Rate (MR$_K$), and Brier-minimum Final Displacement Error (b-minFDE$_K$), where \textit{K} represents the number of predicted (output) trajectories, set as 6.

In specific, minADE$_K$ denotes the average L2 distance between the best-predicted trajectory and the ground truth. The best prediction is the trajectory with the smallest endpoint error among the \textit{K} trajectories. The formula is presented as follows:

\begin{equation}
minADE_{k} =\frac{1}{T'}  {\textstyle \sum_{i=1}^{T'}}  ||P_{i,\hat{k}}^{t} - P_{i,k^{\ast}}^{t}||_{2}  
\end{equation}

minFDE$_K$ represents the L2 distance between the endpoint of the best-predicted trajectory and the ground truth, which can be computed by:

\begin{equation}
minFDE_{k} =||P_{i,\hat{k}}^{T'} - P_{i,k^{\ast}}^{T'}||_{2} 
\end{equation}

Derived from minFDE$_K$, MR$_K$ is defined as the ratio of scenarios where the minFDE$_K$ exceeds two meters. To further enhance the evaluation of uncertainty, $\text{b-minFDE}_K$ incorporates $(1.0 - p)^2$ into the minFDE$_K$, where \textit{p} corresponds to the confidence score of the best-predicted trajectory. 

\textbf{Implementation Details.} The model was trained on an NVIDIA GeForce RTX 4090, employing the Adam optimizer with a batch size of 64. The initial learning rate was set to $5 \times 10^{-4}$, decaying to $1 \times 10^{-4}$ at 45 epochs. Each training session took approximately 80 hours. 
The time required for each trajectory prediction of a single vehicle is approximately 9.7 milliseconds, sufficient to support general online applications. 

\subsection{Determination of the number of neighbors in local attention}

\begin{figure}[htbp]
  \centering
  \begin{minipage}[b]{0.4\textwidth}
    \includegraphics[width=\textwidth]{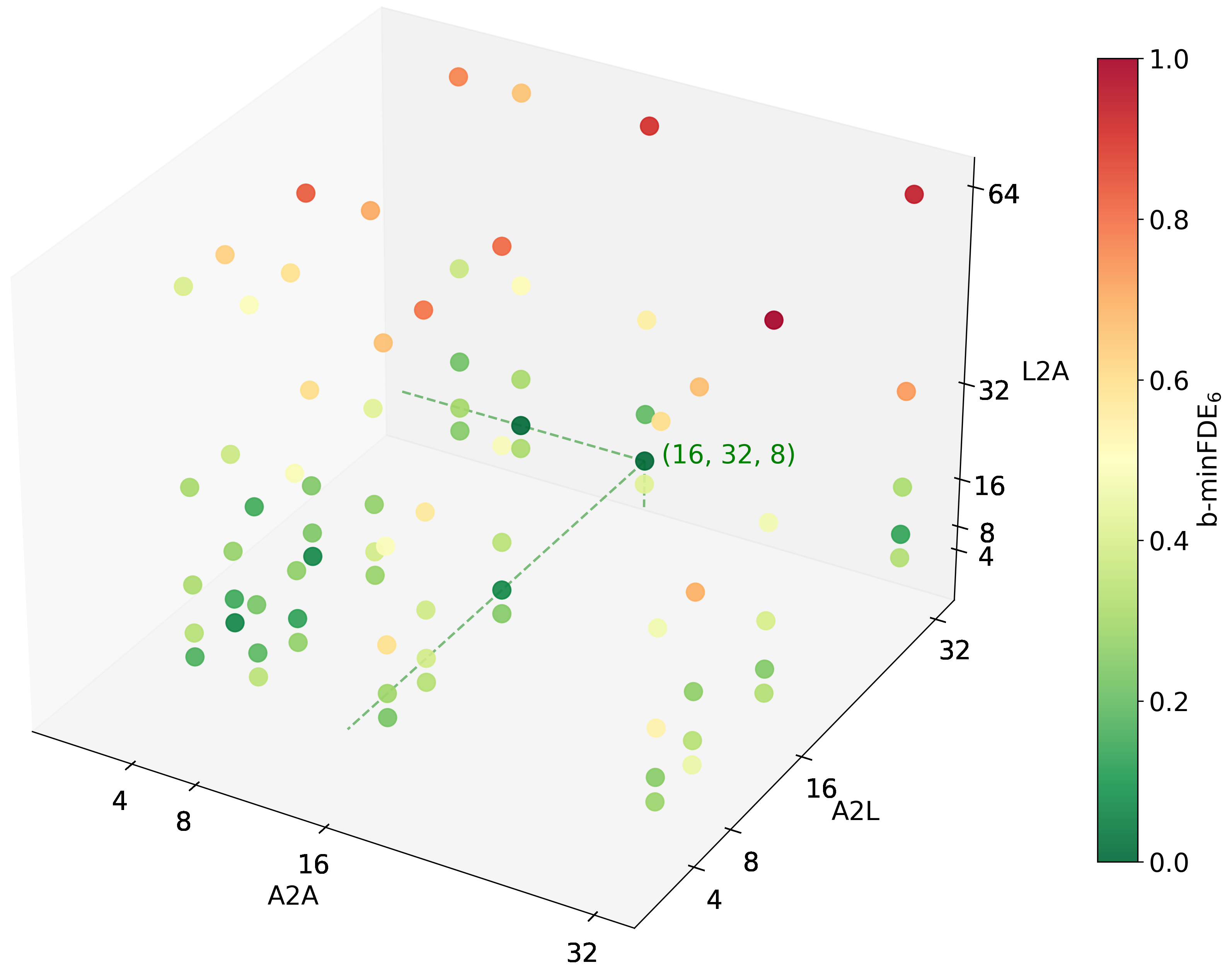}
    \caption{The b-minFDE$_6$ metric with different number of neighbors under the A2A, A2L, and L2A interaction modes.}
    \label{fig:n-number}
  \end{minipage}
\end{figure}

We determine the optimal neighbor numbers for A2A, A2L, and L2A interaction modes via grid search. In these modes, different numbers of neighbors were tested in \autoref{tab:neighbor numbers}, resulting in a total of $4\times4\times5$ combinations. Note that the optimal number of neighbors was determined based on the b-minFDE$_6$ within the first 30 epochs when this metric is nearly stabilized. To better interpret the results, we employ normalization to scale the optimal training results, the results are illustrated in \autoref{fig:n-number}.  

\begin{table}[htbp]
  \centering
  \caption{Number of neighbors for different interaction modes\label{tab:neighbor numbers}}
  \resizebox{0.35\textwidth}{!}{%
    \begin{tabular}{c|ccc}
      \hline
      Interaction mode & A2A & A2L & L2A \\
      \hline
      \multirow{5}{*}{Neighbor numbers} & 4 & 4 & 4 \\
      & 8 & 8 & 8 \\
      & 16 & 16 & 16 \\
      & 32 & 32 & 32 \\
      & - & - & 64 \\
      \hline
    \end{tabular}%
  }
  \label{tab:addlabel}%
\end{table}%

\begin{figure*}[htbp]
  \centering

  \begin{subfigure}[b]{0.325\textwidth}
    \includegraphics[width=\textwidth]{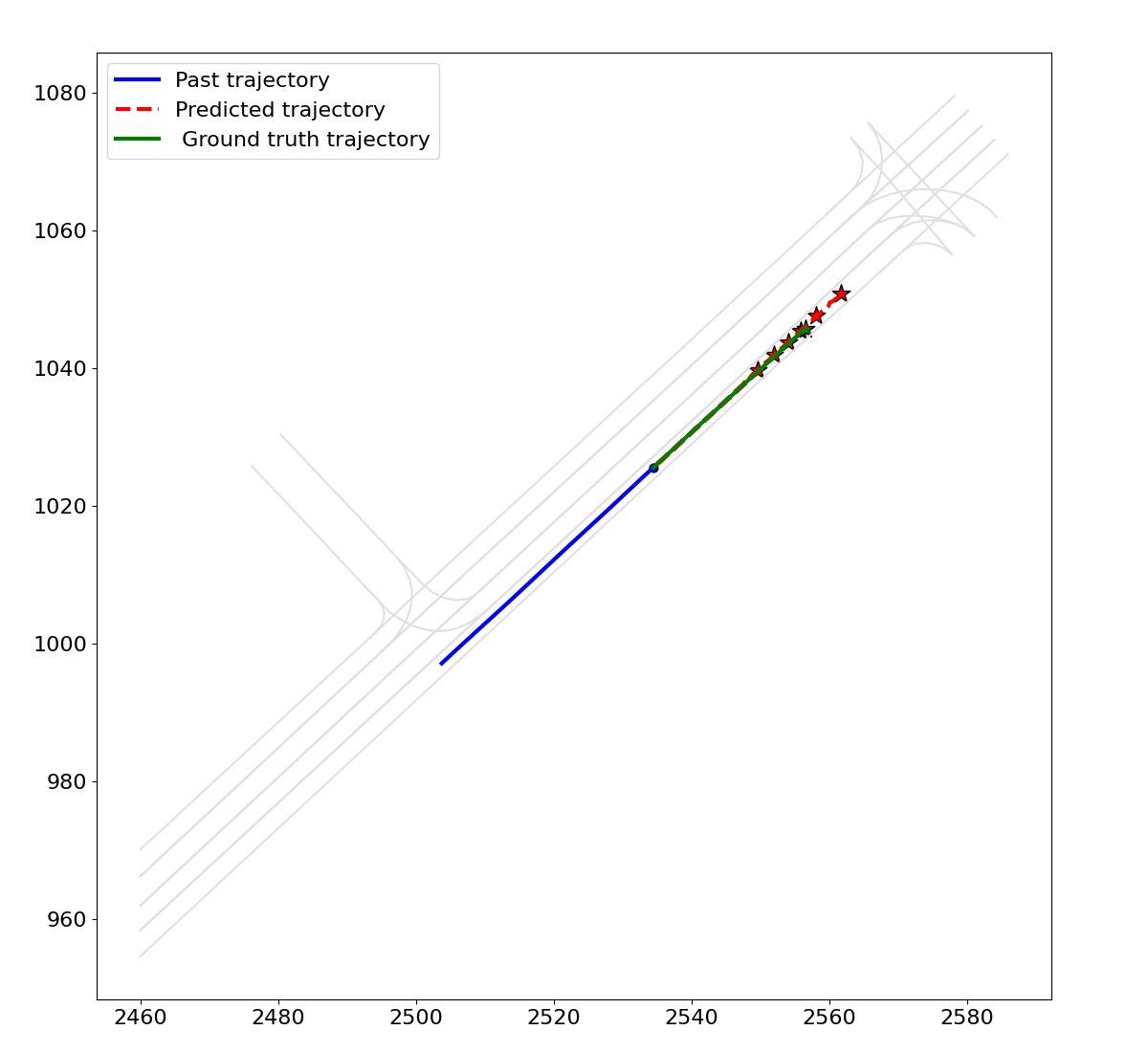}
    \caption{Constant speed}
    \label{fig:a1}
  \end{subfigure}\hspace{0.001\textwidth}
  \begin{subfigure}[b]{0.325\textwidth}
    \includegraphics[width=\textwidth]{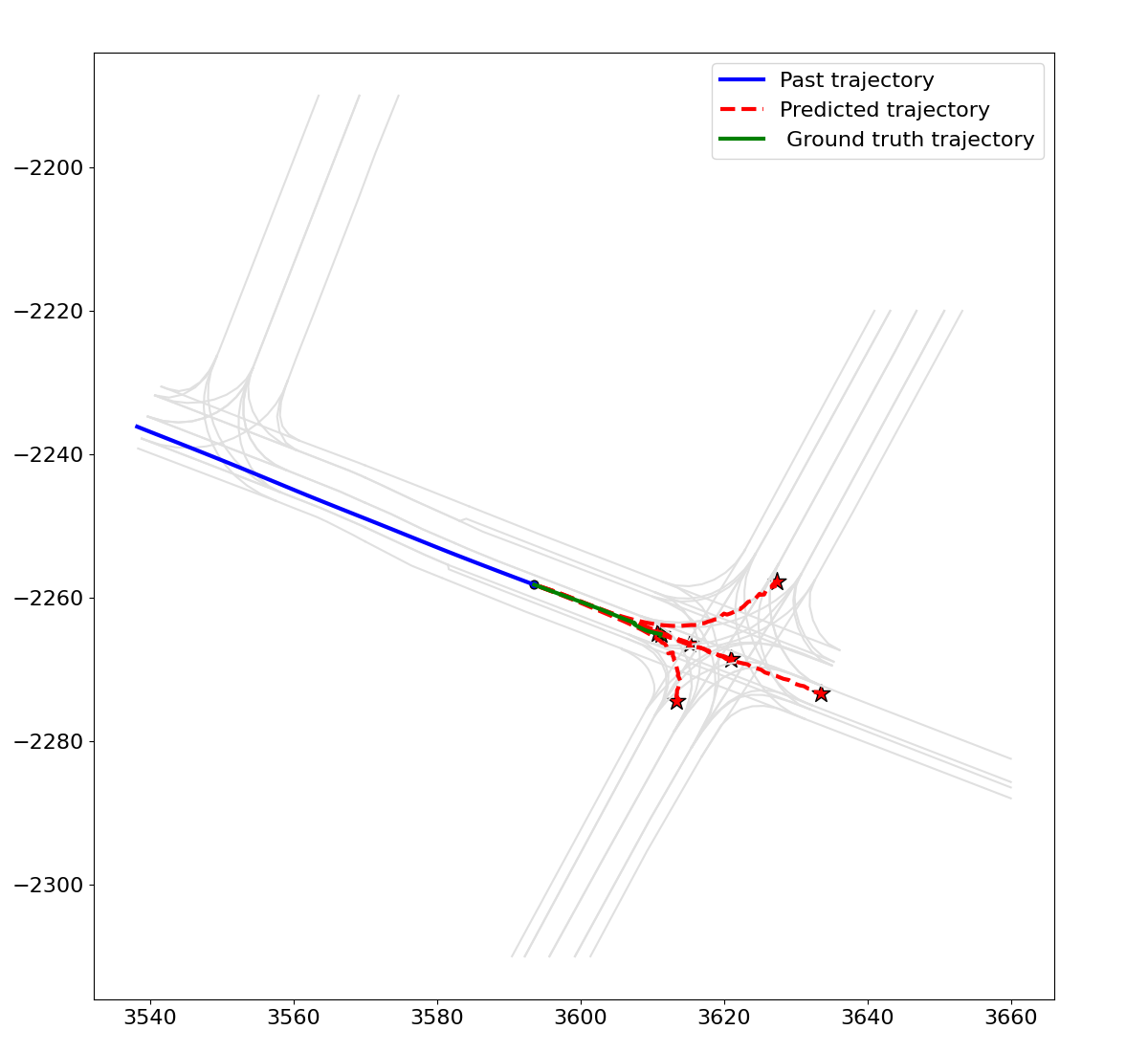}
    \caption{Rapid deceleration}
    \label{fig:a2}
  \end{subfigure}\hspace{0.001\textwidth}
  \begin{subfigure}[b]{0.325\textwidth}
    \includegraphics[width=\textwidth]{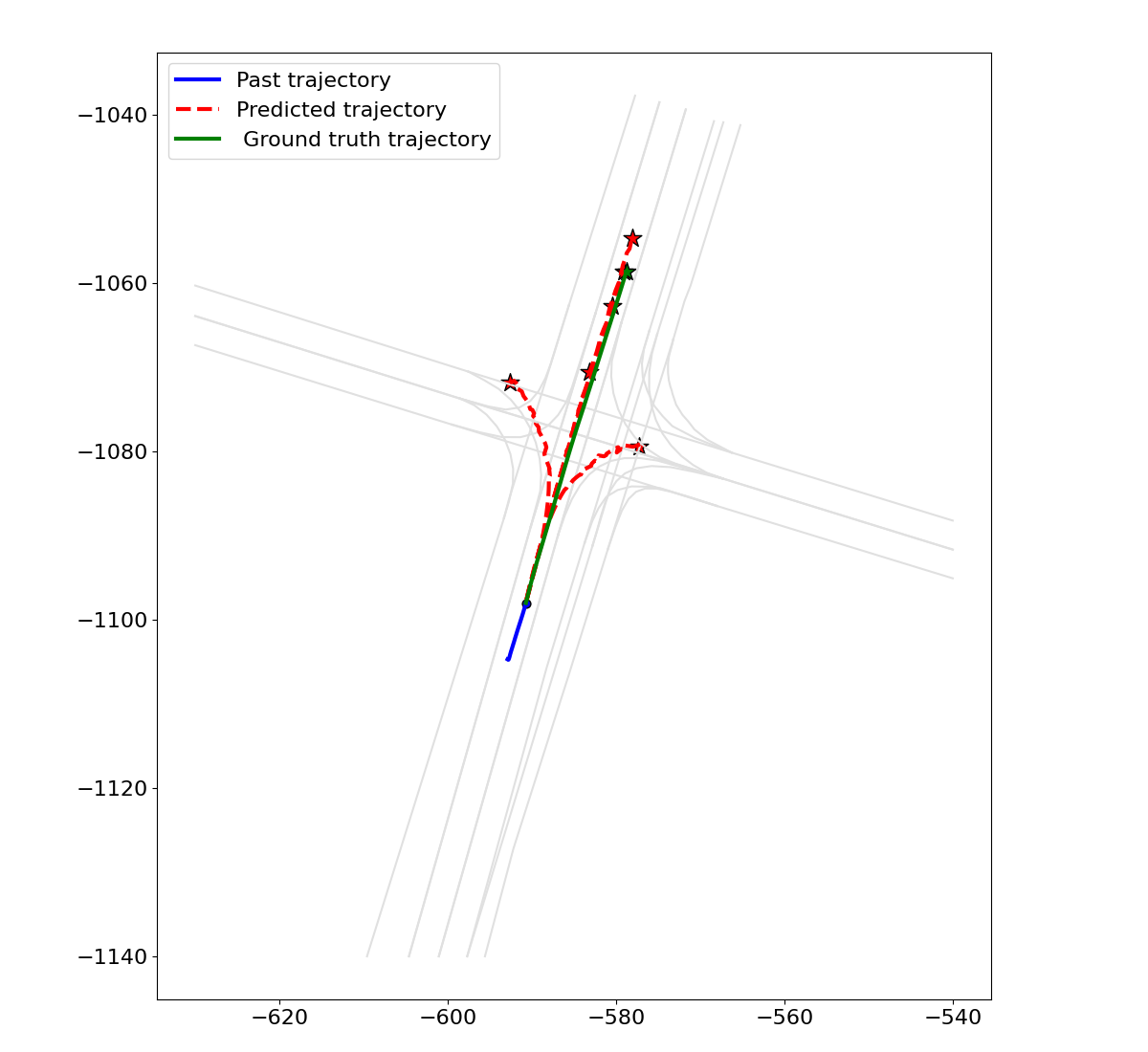}
    \caption{Rapid acceleration}
    \label{fig:a3}
  \end{subfigure}

  \caption{Qualitative results on the Argoverse 2 validation set: w/ embedding map topology matrices and local attention. \label{qualitative1}}
\end{figure*}

\begin{figure*}[htbp]
  \centering
  \begin{subfigure}[b]{0.325\textwidth}
    \includegraphics[width=\textwidth]{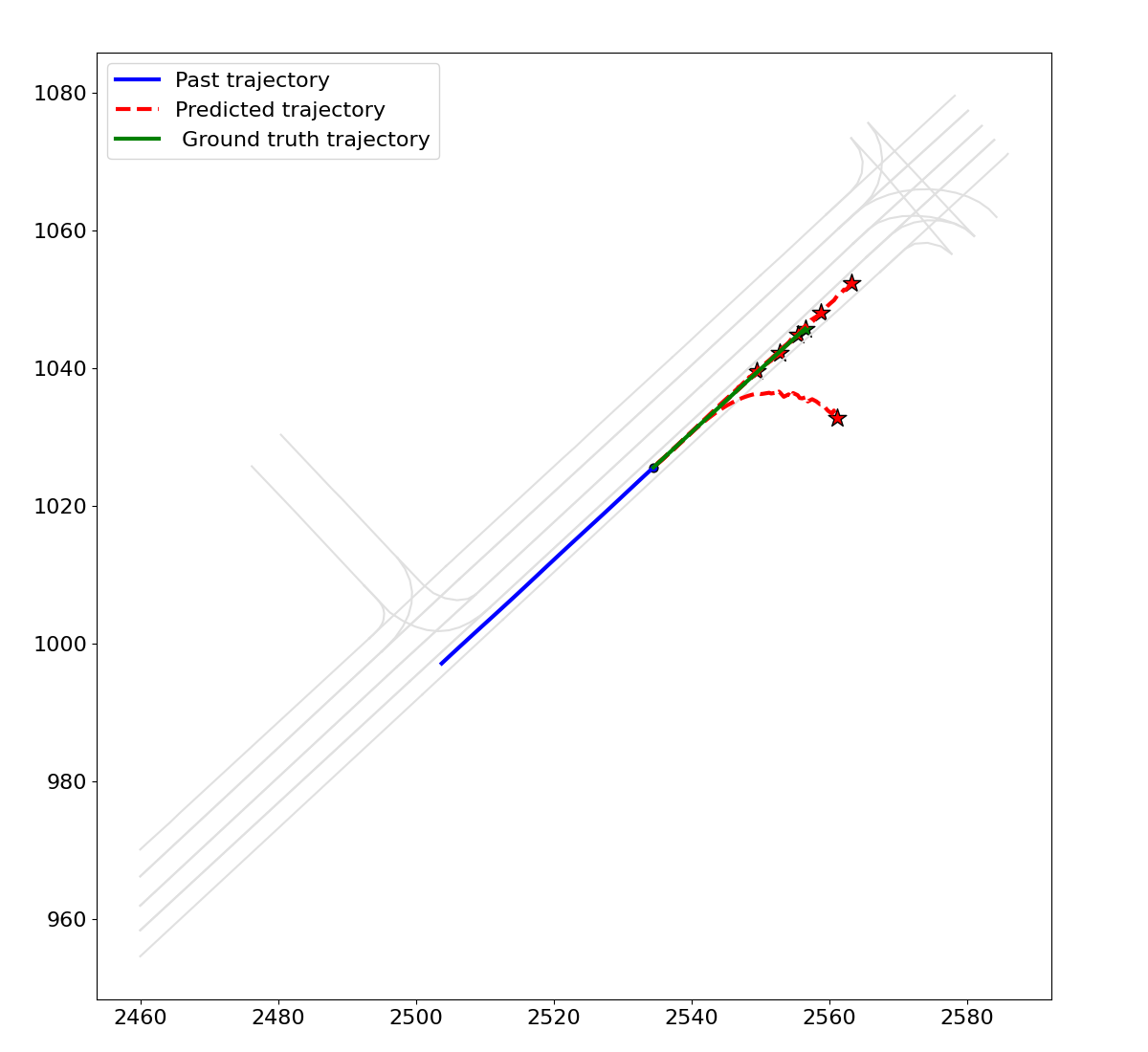}
    \caption{Constant speed}
    \label{fig:b1}
  \end{subfigure}\hspace{0.001\textwidth}
  \begin{subfigure}[b]{0.325\textwidth}
    \includegraphics[width=\textwidth]{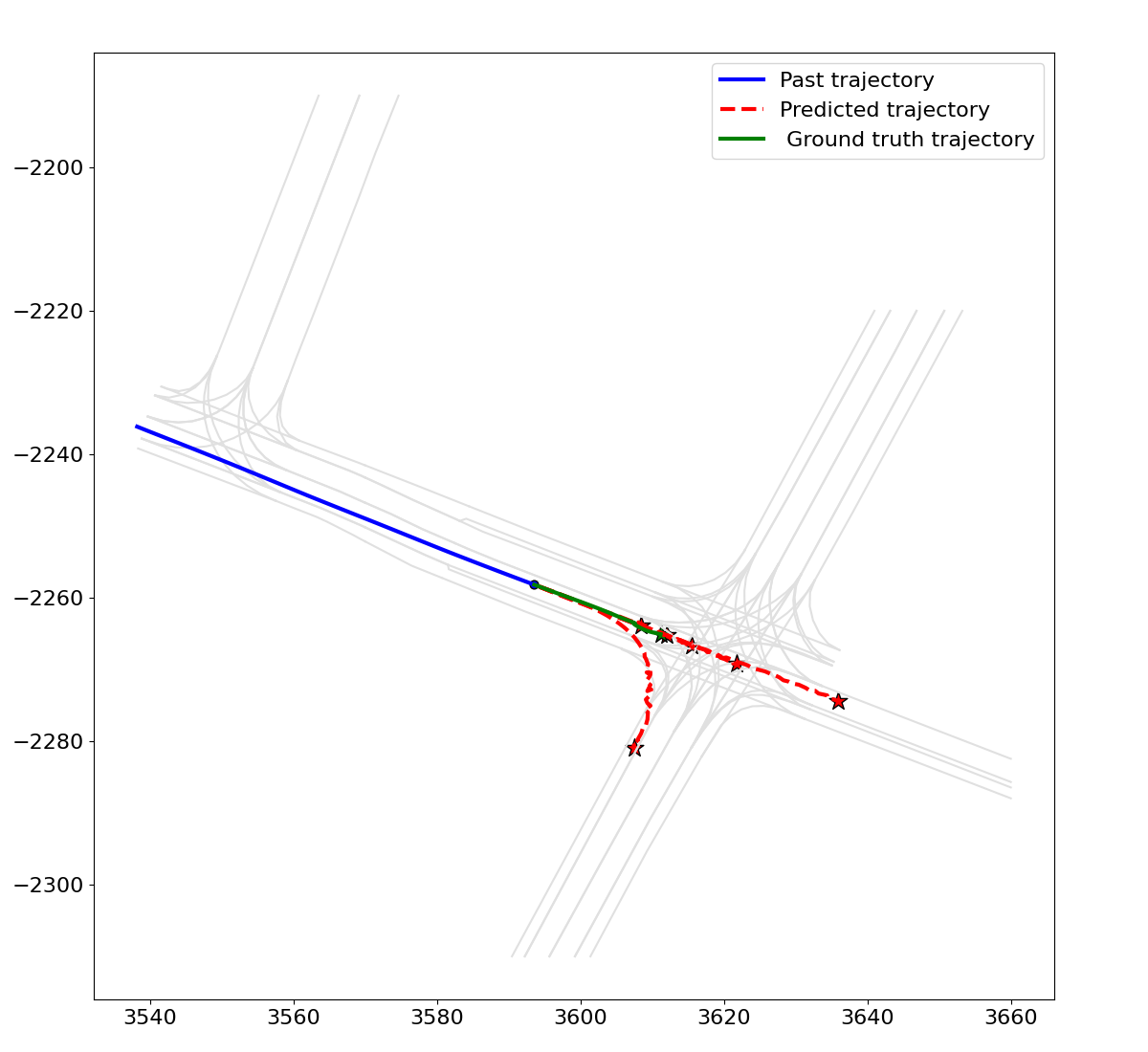}
    \caption{Rapid deceleration}
    \label{fig:b2}
  \end{subfigure}\hspace{0.001\textwidth}
  \begin{subfigure}[b]{0.325\textwidth}
    \includegraphics[width=\textwidth]{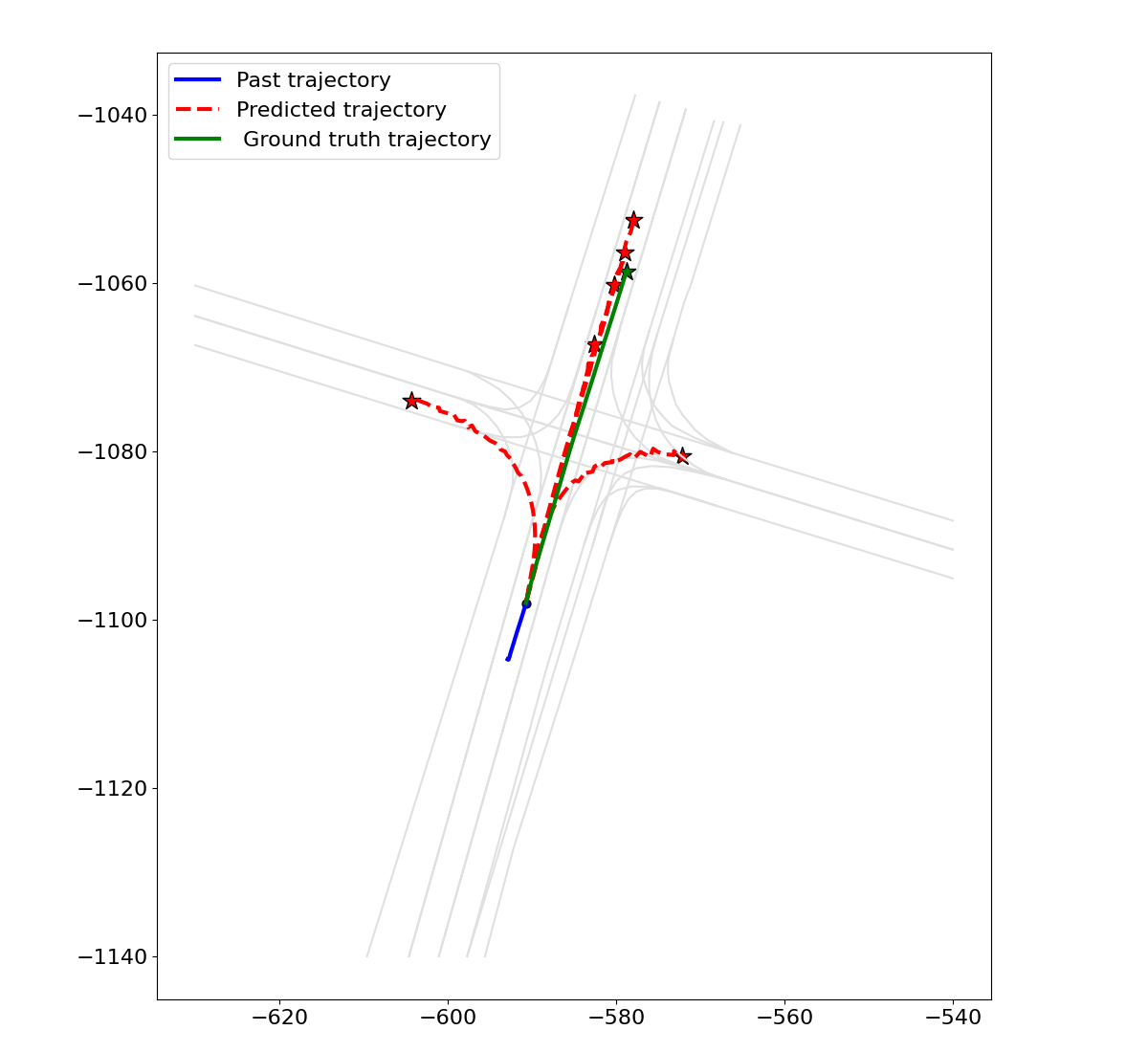}
    \caption{Rapid acceleration}
    \label{fig:b3}
  \end{subfigure}
  \caption{Qualitative results on the Argoverse 2 validation set: w/o embedding map topology matrices and local attention. \label{qualitative2}}
\end{figure*}

For the A2A interaction mode, the b-minFDE$_6$ can achieve better performance when the neighbor value is less than or equal to 16. However, the A2L interaction mode exhibits less sensitivity to surrounding neighbors and achieves the best performance at $e=32$. For the L2A interaction mode, the best performance is achieved at $e=8$, which demonstrates that the agent should focus on the nearest eight lanes. In subsequent experiments, the numbers of neighbors for the A2A, A2L, and L2A modules in the Local Attention part were set as 16, 32, and 8 respectively.

\subsection{Quantitative Results}

In comparisons between the LGT model and the Argoverse 2 baseline models\cite{Argoverse22023} together with some advanced benchmarks, the results are presented in \autoref{tab:quantitatives}. The findings indicate that the proposed model excels in various key metrics, including b-minFDE$_6$, minFDE$_6$, MR$_6$, minADE$_1$, minFDE$_1$, and MR$_1$. Notably, the b-minFDE$_6$ metric, which considers both the accuracy of predicted trajectories and their corresponding probabilities, is usually considered as the most important metric in evaluating model performance~\cite{chang2019argoverse}. A significant reduction in b-minFDE$_6$ is observed for the LGT model compared to other benchmarks, showcasing its superior predictive performance. Compared to the Argoverse 2 baseline (NN), the LGT model has demonstrated substantial improvements across various metrics, with a 60.73\% decrease in minFDE6. In addition, LGT outperforms LaneGCN is five of the total six performance metrics, especially a 2.65\% reduction in the b-minFDE$_6$, the LaneGCN model is often considered as a milestone in vectorized modeling for trajectory prediction. 


\subsection{Ablation Study}
The ablation experiments aim to validate the effectiveness of introducing the bias matrices and the local attention mechanism in enhancing the model's performance. The results in \autoref{tab:ablation} indicate that the introduction of each component has a significant positive impact on the model's predictive accuracy.

In specific, the introduction of matrices \textit{B} is to capture the relative positional information of the map topology structure. This design leads to a crucial contribution to the overall improvement of the model's performance, with a decrease in the b-minFDE$_6$ from 2.83 to 2.74 (about 3.2\% improvement).

Secondly, matrices $D_{inter}$ and $D_{outer}$, representing the SPD encoding, effectively reflect the global information of the map topology structure. These matrices can significantly enhance the prediction accuracy (by 1.1\% reduction in b-minFDE$_6$), highlighting the indispensable role of global information in the model.

Furthermore, the incorporation of the local attention mechanism enables the agent to focus on information from its surrounding neighbors. This add-on plays a key role in improving the model's performance (by 5.54\% reduction in b-minFDE$_6$).

\subsection{Qualitative Results}

Due to the absence of real trajectories in the test set, we have chosen to present qualitative results using the validation set. In \autoref{qualitative1} and \autoref{qualitative2}, we present the predictive outcomes for three challenging scenarios from the Argoverse 2 validation set. In these figures, the blue solid, green solid, and red dashed lines denote the historical observed, ground-truth, and predicted trajectories, respectively. These two sets of images correspond to scenarios with embedded map topology matrices and local attention mechanisms, as well as scenarios without embedding.

The results demonstrate that the integration of the map topology matrices and local attention significantly enhances the consistency between predicted trajectories and the map topology information, leading to an improvement in prediction accuracy. Specifically, in the constant-speed scenario, \autoref{fig:b1} inaccurately evaluates the map structure, predicting a right-turning trajectory against the map topology structure, while \autoref{fig:a1} demonstrates superior prediction results. In the rapid deceleration scenario, compared to \autoref{fig:b2}, \autoref{fig:a2} better identifies map topology structure information, accurately predicting the left-turning trajectory and providing a more precise prediction of the driver's imminent deceleration behavior. In the rapid acceleration scenario, compared to \autoref{fig:b3}, \autoref{fig:a3} exhibits a smoother trajectory with minor prediction errors.

\section{Conclusions}
\label{sec:conclusion}
This paper designed a Lane Graph Transformer (LGT) framework with structure-aware capabilities, applied to lane graph representation for vehicle trajectory prediction. Drawing inspiration from graph theory, the LGT model embedded the map topology information from the HD map in the form of RPE and SPD matrices into the attention structure, effectively identifying the features of the map structure. Compared to representative trajectory prediction models like LaneGCN, the LGT model predicts trajectories on a lane basis, maintaining a high accuracy level for the Argoverse 2 dataset. Another logical extension could lie in predicting the trajectories of multiple traffic participants (including surrounding vehicles, bicycles, and pedestrians) instead of a single vehicle. Efforts will also be made on refining the overall trajectory prediction framework and improving model transferability, facilitating its deployment in real-world traffic applications. In addition, to prioritize efficiency and safety, we will delve deeper into exploring the synergistic relationship between trajectory prediction and planning \& control tasks of autonomous vehicles. 


\bibliographystyle{Transactions-Bibliography/IEEEtran.bst}
\bibliography{ref}

\begin{thebibliography}{10}
\providecommand{\url}[1]{#1}
\csname url@samestyle\endcsname
\providecommand{\newblock}{\relax}
\providecommand{\bibinfo}[2]{#2}
\providecommand{\BIBentrySTDinterwordspacing}{\spaceskip=0pt\relax}
\providecommand{\BIBentryALTinterwordstretchfactor}{4}
\providecommand{\BIBentryALTinterwordspacing}{\spaceskip=\fontdimen2\font plus
\BIBentryALTinterwordstretchfactor\fontdimen3\font minus \fontdimen4\font\relax}
\providecommand{\BIBforeignlanguage}[2]{{%
\expandafter\ifx\csname l@#1\endcsname\relax
\typeout{** WARNING: IEEEtran.bst: No hyphenation pattern has been}%
\typeout{** loaded for the language `#1'. Using the pattern for}%
\typeout{** the default language instead.}%
\else
\language=\csname l@#1\endcsname
\fi
#2}}
\providecommand{\BIBdecl}{\relax}
\BIBdecl

\bibitem{molakis2017}
D.~Milakis, B.~Arem, and B.~Wee, ``Policy and society related implications of automated driving: A review of literature and directions for future research,'' \emph{Journal of Intelligent Transportation Systems Technology Planning and Operations}, vol.~21, pp. 324--348, 02 2017.

\bibitem{QCnet2023}
Z.~Zhou, J.~Wang, Y.~Li, and Y.~Huang, ``Query-centric trajectory prediction,'' in \emph{2023 IEEE/CVF Conference on Computer Vision and Pattern Recognition (CVPR)}, 2023, pp. 17\,863--17\,873.

\bibitem{M2I2022}
Q.~Sun, X.~Huang, J.~Gu, B.~C. Williams, and H.~Zhao, ``M2i: From factored marginal trajectory prediction to interactive prediction,'' in \emph{2022 IEEE/CVF Conference on Computer Vision and Pattern Recognition (CVPR)}, 2022, pp. 6533--6542.

\bibitem{tiv-Survey2022}
Y.~Huang, J.~Du, Z.~Yang, Z.~Zhou, L.~Zhang, and H.~Chen, ``A survey on trajectory-prediction methods for autonomous driving,'' \emph{IEEE Transactions on Intelligent Vehicles}, vol.~7, no.~3, pp. 652--674, 2022.

\bibitem{xie2019mining}
K.~Xie, K.~Ozbay, H.~Yang, and C.~Li, ``Mining automatically extracted vehicle trajectory data for proactive safety analytics,'' \emph{Transportation Research Part C: Emerging Technologies}, vol. 106, pp. 61--72, 2019.

\bibitem{wang2024assessing}
Y.~Wang, C.~Xu, P.~Liu, Z.~Li, and K.~Chen, ``Assessing the predictability of surrogate safety measures as crash precursors based on vehicle trajectory data prior to crashes,'' \emph{Accident Analysis \& Prevention}, vol. 201, p. 107573, 2024.

\bibitem{GNET2023}
X.~Gao, X.~Jia, Y.~Li, and H.~Xiong, ``Dynamic scenario representation learning for motion forecasting with heterogeneous graph convolutional recurrent networks,'' \emph{IEEE Robotics and Automation Letters}, vol.~8, no.~5, pp. 2946--2953, 2023.

\bibitem{IPCC-TP2023}
D.~Zhu, G.~Zhai, Y.~Di, F.~Manhardt, H.~Berkemeyer, T.~Tran, N.~Navab, F.~Tombari, and B.~Busam, ``Ipcc-tp: Utilizing incremental pearson correlation coefficient for joint multi-agent trajectory prediction,'' in \emph{2023 IEEE/CVF Conference on Computer Vision and Pattern Recognition (CVPR)}, 2023, pp. 5507--5516.

\bibitem{laformer2023}
M.~Liu, H.~Cheng, L.~Chen, H.~Broszio, J.~Li, R.~Zhao, M.~Sester, and M.~Y. Yang, ``Laformer: Trajectory prediction for autonomous driving with lane-aware scene constraints,'' 2023.

\bibitem{SocialLSTM2016}
A.~Alahi, K.~Goel, V.~Ramanathan, A.~Robicquet, L.~Fei-Fei, and S.~Savarese, ``Social lstm: Human trajectory prediction in crowded spaces,'' in \emph{2016 IEEE Conference on Computer Vision and Pattern Recognition (CVPR)}, 2016, pp. 961--971.

\bibitem{SS-LSTM2018}
H.~Xue, D.~Q. Huynh, and M.~Reynolds, ``Ss-lstm: A hierarchical lstm model for pedestrian trajectory prediction,'' in \emph{2018 IEEE Winter Conference on Applications of Computer Vision (WACV)}, 2018, pp. 1186--1194.

\bibitem{Social-Attention2018}
A.~Vemula, K.~Muelling, and J.~Oh, ``Social attention: Modeling attention in human crowds,'' in \emph{2018 IEEE International Conference on Robotics and Automation (ICRA)}, 2018, pp. 4601--4607.

\bibitem{tiv-mo2023}
X.~Mo and C.~Lv, ``Predictive neural motion planner for autonomous driving using graph networks,'' \emph{IEEE Transactions on Intelligent Vehicles}, vol.~8, no.~2, pp. 1983--1993, 2023.

\bibitem{tiv-guo2023}
L.~Guo, C.~Shan, T.~Shi, X.~Li, and F.-Y. Wang, ``A vectorized representation model for trajectory prediction of intelligent vehicles in challenging scenarios,'' \emph{IEEE Transactions on Intelligent Vehicles}, pp. 1--6, 2023.

\bibitem{tiv-Spatio-Temporal-Graph2023}
D.~Xu, X.~Shang, Y.~Liu, H.~Peng, and H.~Li, ``Group vehicle trajectory prediction with global spatio-temporal graph,'' \emph{IEEE Transactions on Intelligent Vehicles}, vol.~8, no.~2, pp. 1219--1229, 2023.

\bibitem{Lane-transformer2023}
Z.~Wang, J.~Guo, Z.~Hu, H.~Zhang, J.~Zhang, and J.~Pu, ``Lane transformer: A high-efficiency trajectory prediction model,'' \emph{IEEE Open Journal of Intelligent Transportation Systems}, vol.~4, pp. 2--13, 2023.

\bibitem{tiv-zuo2023}
Z.~Zuo, X.~Wang, S.~Guo, Z.~Liu, Z.~Li, and Y.~Wang, ``Trajectory prediction network of autonomous vehicles with fusion of historical interactive features,'' \emph{IEEE Transactions on Intelligent Vehicles}, pp. 1--13, 2023.

\bibitem{tiv-Interaction-Aware2023}
Z.~Li, Y.~Wang, and Z.~Zuo, ``Interaction-aware prediction for cut-in trajectories with limited observable neighboring vehicles,'' \emph{IEEE Transactions on Intelligent Vehicles}, vol.~8, no.~3, pp. 2148--2161, 2023.

\bibitem{tivAI-TP2023}
K.~Zhang, L.~Zhao, C.~Dong, L.~Wu, and L.~Zheng, ``Ai-tp: Attention-based interaction-aware trajectory prediction for autonomous driving,'' \emph{IEEE Transactions on Intelligent Vehicles}, vol.~8, no.~1, pp. 73--83, 2023.

\bibitem{wang2023wsip}
R.~Wang, S.~Wang, H.~Yan, and X.~Wang, ``Wsip: Wave superposition inspired pooling for dynamic interactions-aware trajectory prediction,'' in \emph{Proceedings of the AAAI Conference on Artificial Intelligence}, vol.~37, no.~4, 2023, pp. 4685--4692.

\bibitem{Rethinking-Graph2021}
D.~Kreuzer, D.~Beaini, W.~L. Hamilton, V.~Létourneau, and P.~Tossou, ``Rethinking graph transformers with spectral attention,'' 2021.

\bibitem{attention2023}
A.~Vaswani, N.~Shazeer, N.~Parmar, J.~Uszkoreit, L.~Jones, A.~N. Gomez, L.~Kaiser, and I.~Polosukhin, ``Attention is all you need,'' 2023.

\bibitem{HiVT2022}
Z.~Zhou, L.~Ye, J.~Wang, K.~Wu, and K.~Lu, ``Hivt: Hierarchical vector transformer for multi-agent motion prediction,'' in \emph{2022 IEEE/CVF Conference on Computer Vision and Pattern Recognition (CVPR)}, 2022, pp. 8813--8823.

\bibitem{Scene-Transformer2022}
J.~Ngiam, B.~Caine, V.~Vasudevan, Z.~Zhang, H.-T.~L. Chiang, J.~Ling, R.~Roelofs, A.~Bewley, C.~Liu, A.~Venugopal, D.~Weiss, B.~Sapp, Z.~Chen, and J.~Shlens, ``Scene transformer: A unified architecture for predicting multiple agent trajectories,'' 2022.

\bibitem{Wayformer2023}
N.~Nayakanti, R.~Al-Rfou, A.~Zhou, K.~Goel, K.~S. Refaat, and B.~Sapp, ``Wayformer: Motion forecasting via simple \& efficient attention networks,'' in \emph{2023 IEEE International Conference on Robotics and Automation (ICRA)}, 2023, pp. 2980--2987.

\bibitem{AgentFormer2021}
Y.~Yuan, X.~Weng, Y.~Ou, and K.~Kitani, ``Agentformer: Agent-aware transformers for socio-temporal multi-agent forecasting,'' in \emph{2021 IEEE/CVF International Conference on Computer Vision (ICCV)}, 2021, pp. 9793--9803.

\bibitem{Directed-Graphs2023}
S.~Geisler, Y.~Li, D.~Mankowitz, A.~T. Cemgil, S.~Günnemann, and C.~Paduraru, ``Transformers meet directed graphs,'' 2023.

\bibitem{Bad-for-Graph2021}
C.~Ying, T.~Cai, S.~Luo, S.~Zheng, G.~Ke, D.~He, Y.~Shen, and T.-Y. Liu, ``Do transformers really perform bad for graph representation?'' 2021.

\bibitem{Not-be-Powerful2022}
S.~Luo, S.~Li, S.~Zheng, T.-Y. Liu, L.~Wang, and D.~He, ``Your transformer may not be as powerful as you expect,'' 2022.

\bibitem{Transformer-for-Graphs2022}
E.~Min, R.~Chen, Y.~Bian, T.~Xu, K.~Zhao, W.~Huang, P.~Zhao, J.~Huang, S.~Ananiadou, and Y.~Rong, ``Transformer for graphs: An overview from architecture perspective,'' 2022.

\bibitem{Biconnectivity2023}
B.~Zhang, S.~Luo, L.~Wang, and D.~He, ``Rethinking the expressive power of gnns via graph biconnectivity,'' 2023.

\bibitem{LaneGCN2020}
M.~Liang, B.~Yang, R.~Hu, Y.~Chen, R.~Liao, S.~Feng, and R.~Urtasun, ``Learning lane graph representations for motion forecasting,'' 2020.

\bibitem{VectorNet2020}
J.~Gao, C.~Sun, H.~Zhao, Y.~Shen, D.~Anguelov, C.~Li, and C.~Schmid, ``Vectornet: Encoding hd maps and agent dynamics from vectorized representation,'' in \emph{2020 IEEE/CVF Conference on Computer Vision and Pattern Recognition (CVPR)}, 2020, pp. 11\,522--11\,530.

\bibitem{Mo2022}
X.~Mo, Z.~Huang, Y.~Xing, and C.~Lv, ``Multi-agent trajectory prediction with heterogeneous edge-enhanced graph attention network,'' \emph{IEEE Transactions on Intelligent Transportation Systems}, vol.~23, no.~7, pp. 9554--9567, 2022.

\bibitem{Multimodal2021}
Y.~Liu, J.~Zhang, L.~Fang, Q.~Jiang, and B.~Zhou, ``Multimodal motion prediction with stacked transformers,'' in \emph{2021 IEEE/CVF Conference on Computer Vision and Pattern Recognition (CVPR)}, 2021, pp. 7573--7582.

\bibitem{Multi-modal2021}
Z.~Huang, X.~Mo, and C.~Lv, ``Multi-modal motion prediction with transformer-based neural network for autonomous driving,'' 2021.

\bibitem{Multi-Head2020}
J.~Mercat, T.~Gilles, N.~El~Zoghby, G.~Sandou, D.~Beauvois, and G.~P. Gil, ``Multi-head attention for multi-modal joint vehicle motion forecasting,'' in \emph{2020 IEEE International Conference on Robotics and Automation (ICRA)}, 2020, pp. 9638--9644.

\bibitem{generalization-dwivedi2021}
V.~P. Dwivedi and X.~Bresson, ``A generalization of transformer networks to graphs,'' 2021.

\bibitem{GRPE2022}
W.~Park, W.~Chang, D.~Lee, J.~Kim, and S.~won Hwang, ``Grpe: Relative positional encoding for graph transformer,'' 2022.

\bibitem{DenseTNT2021}
J.~Gu, C.~Sun, and H.~Zhao, ``Densetnt: End-to-end trajectory prediction from dense goal sets,'' in \emph{2021 IEEE/CVF International Conference on Computer Vision (ICCV)}, 2021, pp. 15\,283--15\,292.

\bibitem{TNT2020}
H.~Zhao, J.~Gao, T.~Lan, C.~Sun, B.~Sapp, B.~Varadarajan, Y.~Shen, Y.~Shen, Y.~Chai, C.~Schmid, C.~Li, and D.~Anguelov, ``Tnt: Target-driven trajectory prediction,'' 2020.

\bibitem{MultiPath++2021}
B.~Varadarajan, A.~Hefny, A.~Srivastava, K.~S. Refaat, N.~Nayakanti, A.~Cornman, K.~Chen, B.~Douillard, C.~P. Lam, D.~Anguelov, and B.~Sapp, ``Multipath++: Efficient information fusion and trajectory aggregation for behavior prediction,'' 2021.

\bibitem{Limits-of-Transfer2023}
C.~Raffel, N.~Shazeer, A.~Roberts, K.~Lee, S.~Narang, M.~Matena, Y.~Zhou, W.~Li, and P.~J. Liu, ``Exploring the limits of transfer learning with a unified text-to-text transformer,'' 2023.

\bibitem{selfattentionrpe2018}
P.~Shaw, J.~Uszkoreit, and A.~Vaswani, ``Self-attention with relative position representations,'' 2018.

\bibitem{Global-Intention2023}
S.~Shi, L.~Jiang, D.~Dai, and B.~Schiele, ``Motion transformer with global intention localization and local movement refinement,'' 2023.

\bibitem{GANet2023}
M.~Wang, X.~Zhu, C.~Yu, W.~Li, Y.~Ma, R.~Jin, X.~Ren, D.~Ren, M.~Wang, and W.~Yang, ``Ganet: Goal area network for motion forecasting,'' 2023.

\bibitem{ssllanes2022}
P.~Bhattacharyya, C.~Huang, and K.~Czarnecki, ``Ssl-lanes: Self-supervised learning for motion forecasting in autonomous driving,'' 2022.

\bibitem{tiv-intrinsic-interaction2023}
J.~lian, S.~Li, D.~Yang, J.~Zhang, and L.~Li, ``Encoding the intrinsic interaction information for vehicle trajectory prediction,'' \emph{IEEE Transactions on Intelligent Vehicles}, pp. 1--12, 2023.

\bibitem{leaderboard}
``Argoverse 2: Motion forecasting competition,'' \url{https://eval.ai/web/challenges/challenge-page/1719/overview}.

\bibitem{Argoverse22023}
B.~Wilson, W.~Qi, T.~Agarwal, J.~Lambert, J.~Singh, S.~Khandelwal, B.~Pan, R.~Kumar, A.~Hartnett, J.~K. Pontes, D.~Ramanan, P.~Carr, and J.~Hays, ``Argoverse 2: Next generation datasets for self-driving perception and forecasting,'' 2023.

\bibitem{chang2019argoverse}
M.-F. Chang, J.~Lambert, P.~Sangkloy, J.~Singh, S.~Bak, A.~Hartnett, D.~Wang, P.~Carr, S.~Lucey, D.~Ramanan \emph{et~al.}, ``Argoverse: {3D} tracking and forecasting with rich maps,'' in \emph{Proceedings of the IEEE/CVF Conference on Computer Vision and Pattern Recognition}, 2019, pp. 8748--8757.

\end{thebibliography}


%





\ifCLASSOPTIONcaptionsoff
  \newpage
\fi

\begin{IEEEbiography}[{\includegraphics[width=1in,height=1.25in,clip,keepaspectratio]{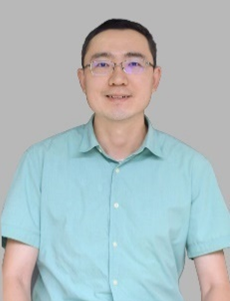}}]{Zhanbo Sun} received his Ph.D. degree from Rensselaer Polytechnic Institute in 2014 and the B.E. degree from Tsinghua University in 2009, respectively. He is now a Professor in the School of Transportation and Logistics at Southwest Jiaotong University. His research focuses on vehicle-road automation and coordination, green transportation, and rail transit. He has won the Jeme Tianyou Railway Science and Technology Award and the Best Paper Award of the International Federation of Automatic Control (IFAC).
\end{IEEEbiography}

\begin{IEEEbiography}[{\includegraphics[width=1in,height=1.25in,clip,keepaspectratio]{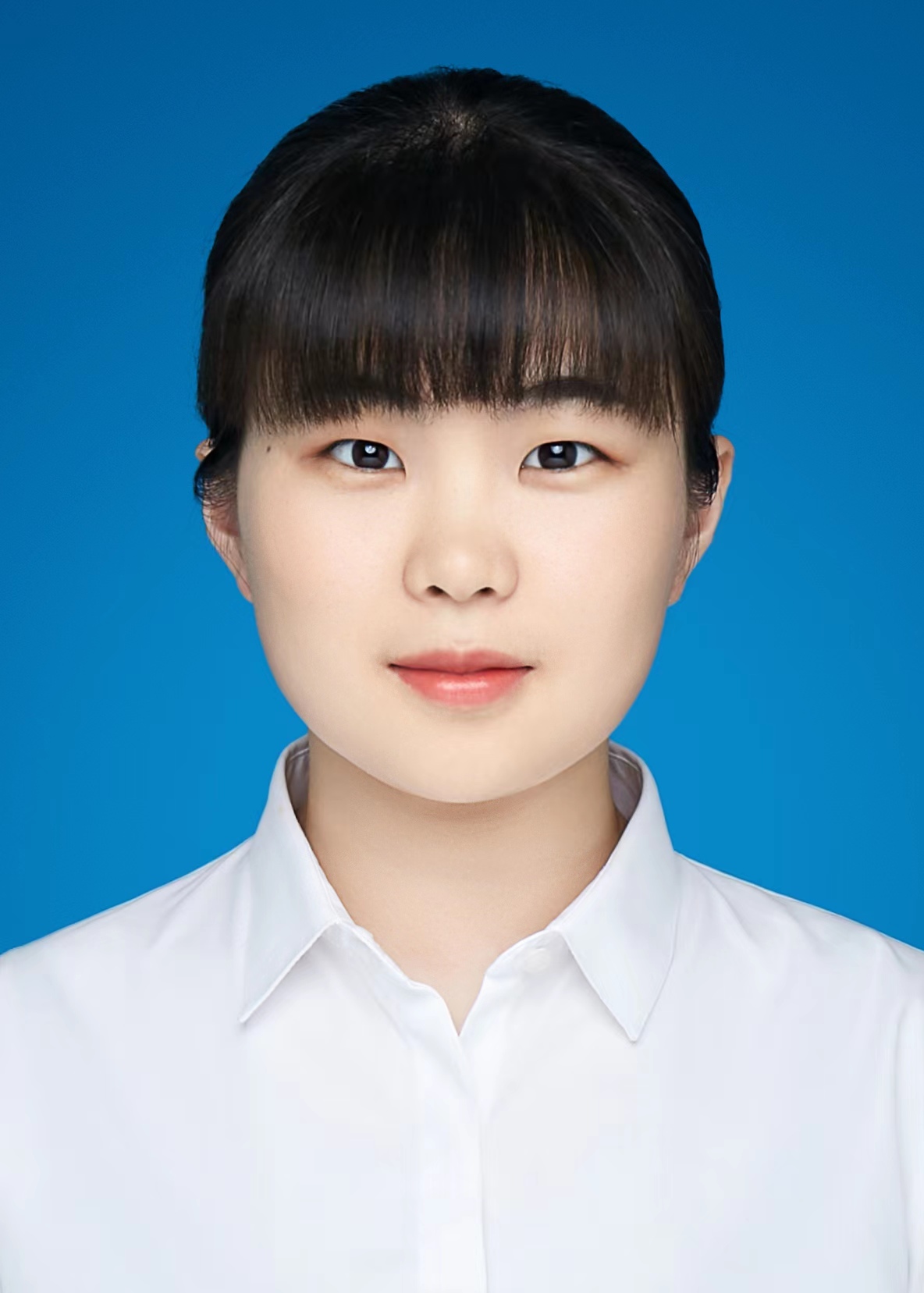}}]{Caiyin Dong} received her B.S degree from Soochow University in 2022. Currently, she is pursuing the M.S. degree with Southwest Jiaotong University. Her research interests include 
 deep learning, trajectory prediction in autonomous driving systems.
\end{IEEEbiography}

\begin{IEEEbiography}[{\includegraphics[width=1in,height=1.25in,clip,keepaspectratio]{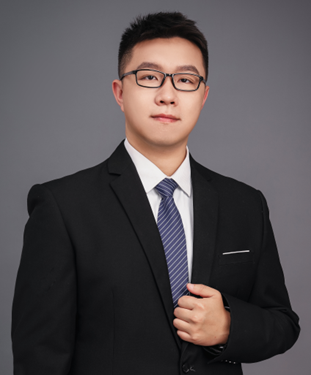}}]{Ang Ji} (M’23) received the Ph.D. degree from the School of Civil Engineering, The University of Sydney, Australia, in 2022. He now serves as an Assistant Professor at the School of Transportation and Logistics, Southwest Jiaotong University, Chengdu, China. His research interests include connected automated vehicles, deep/reinforcement learning, microscopic traffic modeling, and game-theoretic applications.
\end{IEEEbiography}

\begin{IEEEbiography}[{\includegraphics[width=1in,height=1.25in,clip,keepaspectratio]{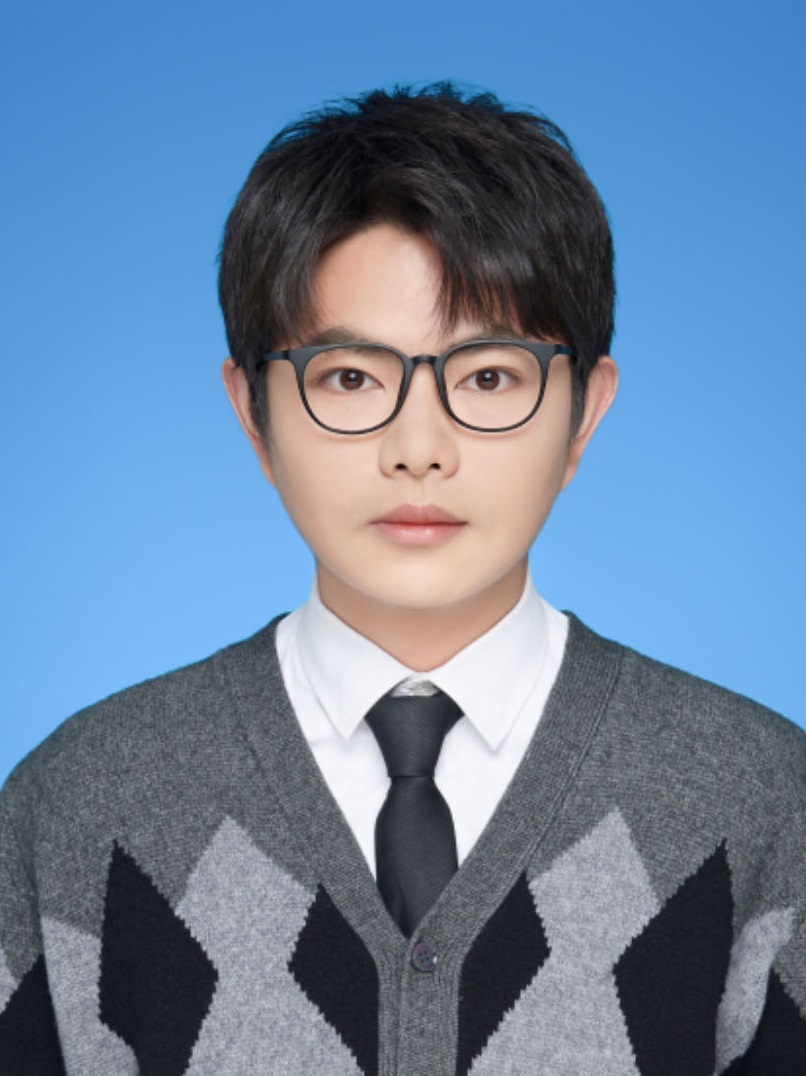}}]{Ruibin Zhao} 
received his M.S degree from Southwest Jiaotong University in 2023. Currently, he works as an algorithm developer in the intelligent autonomous vehicle group at Plus.ai. His research interests focus on trajectory prediction, driving behavior, and motion planning.
\end{IEEEbiography}

\begin{IEEEbiography}[{\includegraphics[width=1in,height=1.25in,clip,keepaspectratio]{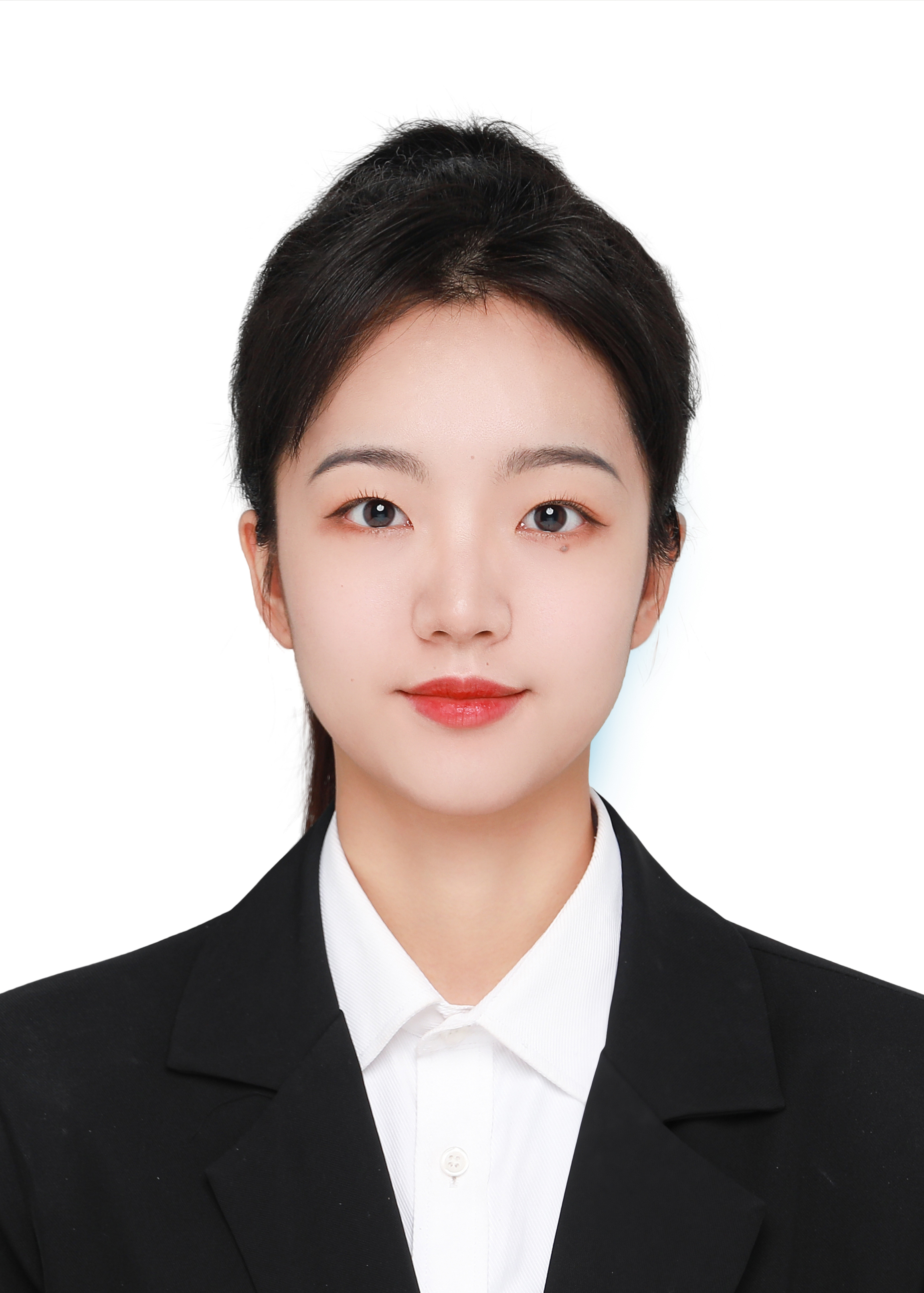}}]{Yu Zhao} received her B.S degree from Southwest Jiaotong University in 2022. Currently, she is pursuing the M.S. degree with Southwest Jiaotong University. Her research interests focus on the application of deep learning in traffic flow forecasting, trajectory planning, and microscopic traffic modeling.
\end{IEEEbiography}





\end{document}